\documentclass[letterpaper]{article} 
\usepackage{aaai2026}  
\usepackage{times}  
\usepackage{helvet}  
\usepackage{courier}  
\usepackage[hyphens]{url}  
\usepackage[section]{placeins}
\usepackage{graphicx} 
\usepackage{natbib}  
\usepackage{caption} 
\usepackage{adjustbox}
\usepackage{amsthm}
\usepackage{amssymb}
\usepackage{dsfont}        
\usepackage{amsmath}
\usepackage{subcaption}
\usepackage{booktabs}
\usepackage{xcolor}
\usepackage{colortbl}
\usepackage{pifont}  
\usepackage{multirow}
\usepackage{float}              
\usepackage{enumitem}           

\usepackage{amsmath,amssymb,amsfonts}

\usepackage{dsfont}          
\let\mathbbm\mathds          

\newtheorem{theorem}{Theorem}
\newtheorem{lemma}{Lemma}

\begin{document}

\newcommand{\cmark}{\ding{51}}%
\newcommand{\xmark}{\ding{55}}%
\newcommand{\circleorange}{\textcolor{orange}{\large\textbullet}}
\newcommand{\circleblue}{\textcolor{blue}{\large\textbullet}}
\newcommand{\circlegreen}{\textcolor{green}{\large\textbullet}}
\newcommand{\circleblack}{\textcolor{black}{\large\textbullet}}
\newcommand{\circlered}{\textcolor{red}{\large\textbullet}}
\newcommand{\circlepink}{\textcolor{pink}{\large\textbullet}}

\title{FAST-CAD: A Fairness-Aware Framework for Non-Contact Stroke Diagnosis}

\author{
  \textbf{Tommy Sha}\\[4pt]
  \textit{Co-second authors:}\\
  Zhan Cheng\textsuperscript{\dag}\quad
  Haotian Zhai\textsuperscript{\dag}\quad
  Xuwei Ding\textsuperscript{\dag}\quad
  Junnan Li\textsuperscript{\dag}\quad
  Haixiang Tang\textsuperscript{\dag}\quad
  Zaoting Sun\textsuperscript{\dag}\\
  Yanchuan Tang\textsuperscript{\dag}\quad
  Yongzhe (Kindred) Yi\textsuperscript{\dag}\quad
  Yuan Gao\textsuperscript{\dag}\quad
  Anhao Li\textsuperscript{\dag}
}

\affiliations{
  \dag~Co-second authors.
}
\maketitle

\begin{abstract}
Stroke is an acute cerebrovascular disease, and timely diagnosis significantly improves patient survival. However, existing automated diagnosis methods suffer from fairness issues across demographic groups, potentially exacerbating healthcare disparities. In this work, we address this critical challenge by proposing FAST-CAD, a theoretically grounded framework that integrates Domain-Adversarial Training (DAT) with Group Distributionally Robust Optimization (Group-DRO) for fair and accurate non-contact stroke diagnosis. Our approach is built upon rigorous theoretical foundations from domain adaptation theory and minimax fairness, providing convergence guarantees and fairness bounds. We establish a comprehensive multimodal dataset generated via Unreal Engine 5 (UE5) photorealistic simulation, encompassing 2,430 subjects across 12 demographic subgroups defined by age, gender, and posture combinations. Blinded clinical evaluation by board-certified neurologists confirms that only 4.7\% of synthetic samples were identified as non-real. FAST-CAD employs self-supervised encoders with adversarial domain discrimination to learn demographic-invariant representations, while Group-DRO ensures robust performance across all subgroups by optimizing worst-group risk. Extensive experiments demonstrate that our method achieves superior diagnostic performance (92.5\% AUC) while maintaining fairness across all demographic groups, with theoretical validation confirming the effectiveness of our unified DAT+Group-DRO framework. Our work provides both practical advances and theoretical insights for fair medical AI systems; the project page is available at \url{https://shatianming5.github.io/Fast-CAD/}.
\end{abstract}

\section{Introduction}
Stroke is an acute cerebrovascular disease, ranking as the second leading cause of death and the third leading cause of disability worldwide~\cite{johnson2016stroke}. Although patient survival rates have improved in recent years, the incidence of permanent disability remains high, primarily because timely diagnosis within the stroke ``golden window'' are often not achieved and because emergency care and rehabilitation knowledge are insufficiently disseminated at the community level. Critically, existing automated diagnosis systems exhibit significant performance disparities across demographic groups—particularly age, gender, and physical conditions—potentially exacerbating healthcare inequities and undermining trust in AI-assisted medical systems.

\begin{figure}[t]
  \centering
  \includegraphics[width=1.0\columnwidth]{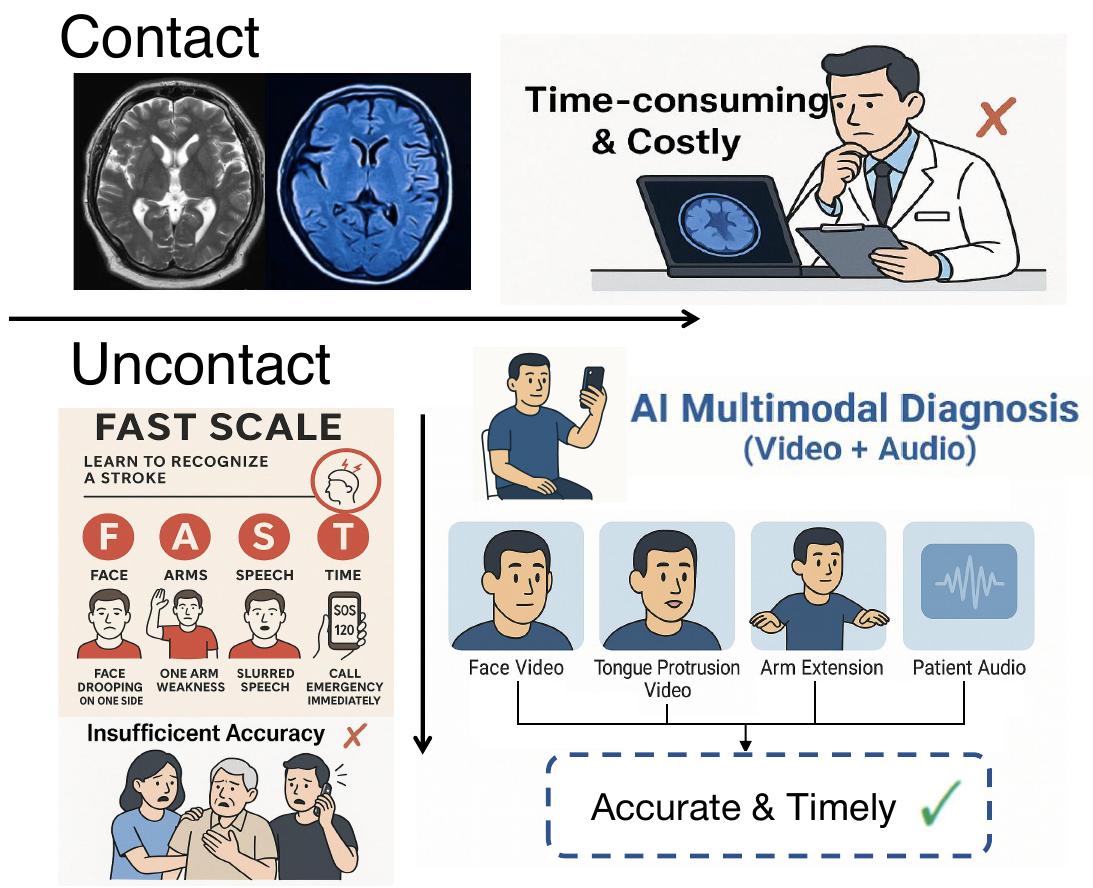}
  \caption{Stroke assessment evolution—from slow, contact-dependent imaging and low-accuracy FAST checks to our novel non‑contact method.}
  \label{fig1}
\end{figure}

Current stroke diagnosis methods can be categorized into contact and non‑contact approaches. Contact methods focus on MRI and CT scan results and are only available in hospitals or specialized medical institutions; they require patients to attend in person and depend on time‑consuming, complex procedural workflows. As a result, untimely diagnosis distances patients from the stroke ``golden window'', compromising the efficacy of subsequent therapeutic interventions. Non‑contact methods (such as the FAST scale~\cite{mohd2004agreement}), while obviating the need for complex workflows and in‑person attendance, require emergency care and rehabilitation knowledge that is insufficiently disseminated at the community level and exhibit reduced accuracy when applied by non‑medical individuals, thereby limiting their widespread applicability. Moreover, both approaches suffer from systematic biases that lead to differential diagnostic accuracy across demographic subgroups, raising critical concerns about algorithmic fairness in healthcare applications.

In order to overcome the aforementioned limitations, researchers have turned their attention to deep learning technology, which has shown immense potential in certain fields\cite{zhai2025mitigating,chen2025multi,cheng2025atom,chen2025tskan}. Current deep learning methods are mainly applied to contact‑based stroke diagnosis, aiming to eliminate the manual process of interpreting MRI and CT scan results, thus achieving automated contact‑based stroke diagnosis.

\textbf{However, do these methods address the critical issues of timely diagnosis and fair treatment for all stroke patients?}

These methods focus on optimizing the extraction of stroke‑related information from MRI and CT scans to enable automated contact‑based stroke diagnosis. However, they address only a small segment of the contact‑based stroke ''golden window''. By the time patients undergo hospital‑based imaging, the utility of automated diagnosis is greatly diminished. More critically, these approaches lack theoretical guarantees for fairness across demographic groups, potentially leading to systematic biases that disproportionately affect vulnerable populations.

\begin{figure}[ht]
  \centering
  \includegraphics[width=1.0\columnwidth]{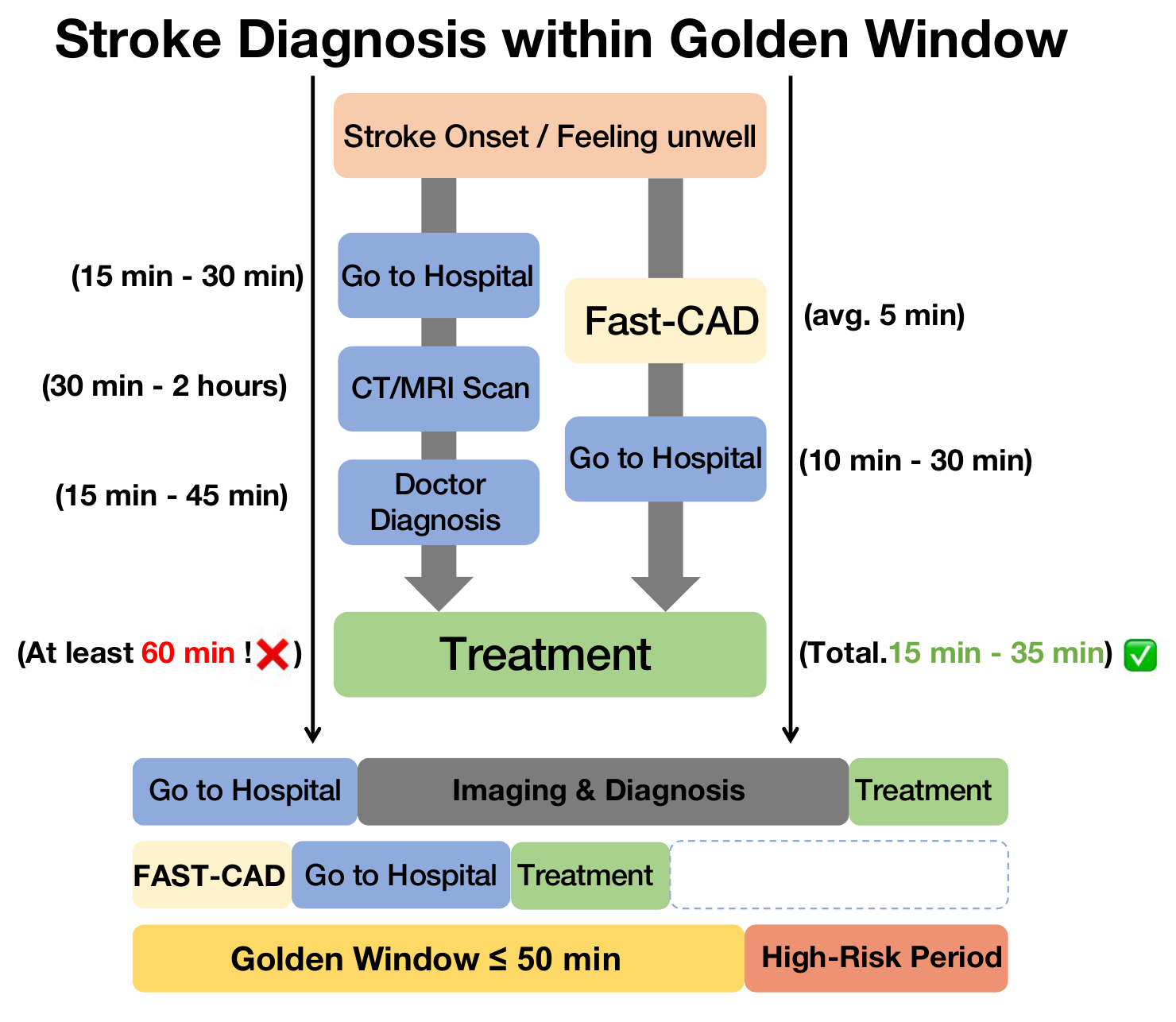}
  \caption{Comparison of the conventional versus FAST-CAD–assisted stroke workflow.}
  \label{fig2}
\end{figure}

Apparently, these methods fail to address the critical issues of timely diagnosis and algorithmic fairness in stroke patients. Hence, we have turned our attention to automated non‑contact stroke diagnosis with theoretically-grounded fairness guarantees, aiming to truly optimize diagnosis within the stroke ''golden window'' while ensuring equitable treatment across all demographic groups.


However, deploying non-contact stroke diagnosis in real-world settings faces three critical challenges that existing methods fail to address: (i) \textbf{Complex environmental conditions:} Emergency rooms and community settings present noisy, uncontrolled environments with varying lighting, background sounds, and distractions that severely degrade diagnostic accuracy; (ii) \textbf{Diverse patient postures:} Stroke patients may present in various postures—sitting upright, lying in bed, or collapsed—each affecting the manifestation of symptoms and requiring adaptive diagnostic approaches; and (iii) \textbf{Demographic disparities:} Current systems exhibit significant performance gaps across age, gender, and physical conditions, with elderly females in sleeping postures experiencing up to 15\% lower diagnostic accuracy, raising serious concerns about healthcare equity.

These challenges fundamentally stem from the lack of comprehensive datasets that capture real-world diagnostic complexity and the absence of theoretically-grounded methods that guarantee fairness. To address this gap, we constructed a large-scale multimodal audiovisual dataset specifically designed for fair non‑contact stroke diagnosis using Unreal Engine 5 (UE5) photorealistic simulation. Our dataset encompasses 2,430 subjects across 12 carefully balanced demographic subgroups, generated over a two-month production period with clinically validated realism.

Building on this dataset, we introduce FAST-CAD, a theoretically-grounded framework that synergistically combines Domain-Adversarial Training (DAT) with Group Distributionally Robust Optimization (Group-DRO). Our key insight: achieving both high accuracy and fairness requires learning representations that are discriminative for stroke detection yet invariant to demographic attributes.

FAST-CAD integrates three components: (i) \textbf{Domain-adversarial learning} that prevents encoders from predicting demographic attributes through gradient reversal; (ii) \textbf{Group-DRO} that explicitly optimizes worst-group performance rather than average metrics; and (iii) \textbf{Self-supervised multimodal fusion} using pretrained encoders (SeCo~\cite{yao2021seco}, HuBERT~\cite{9585401}) with our novel Alternating Dual-Stream Transformer.

Extensive experiments validate our approach: FAST-CAD achieves 92.5\% AUC while reducing the fairness gap by 63\%, maintaining worst-group performance within 1.7\% of average. External validation on 860 subjects under different conditions confirms strong generalization. We contribute: (i) A unified DAT+Group-DRO framework with theoretical guarantees; (ii) A large-scale demographically-stratified dataset (2,430 subjects, 12 balanced subgroups) generated via UE5 simulation; and (iii) The FAST-CAD architecture achieving state-of-the-art performance with fairness guarantees\footnote{Code is provided in the Supplementary Material.}.

\section{Related Works}


\subsection{Stroke Diagnosis}
Contact-based methods (CT~\cite{Sanelli1045}, MRI) provide accurate diagnosis but miss the critical "golden window," while non‑contact approaches (FAST score~\cite{mohd2004agreement}) enable pre-hospital screening but lack accuracy. Deep learning on medical imaging~\cite{Shinohara2019,Lisowska2017ContextAwareCN} accelerates expert diagnosis but remains hospital-bound. Recent non‑contact methods~\cite{CAI2022102522,cai2024m3stroke} achieve 80.85\% accuracy but require controlled postures and lack fairness guarantees across demographic groups, motivating our theoretically-grounded fair diagnosis framework.

\subsection{Self-supervised Learning}
Self-supervised learning enables effective representation learning from limited labeled data. In vision, contrastive methods (MoCo~\cite{9157636}, SimCLR~\cite{chen2020simpleframeworkcontrastivelearning}) learn transferable features without annotation. SeCo~\cite{yao2021seco} extends this to video via spatio-temporal InfoNCE loss. In speech, HuBERT~\cite{9585401} discovers phonetic structures through masked prediction. We leverage these advances for stroke-relevant feature extraction from limited medical data.

\subsection{Domain-Adversarial Training and Fairness in ML}
Domain-adversarial training, pioneered by Ganin and Lempitsky~\cite{ganin2015unsupervised}, addresses domain shift by learning representations that are invariant to domain-specific variations. The theoretical foundation lies in Ben-David et al.'s domain adaptation theory~\cite{ben2010theory}, which establishes upper bounds on target domain error in terms of source domain error and domain discrepancy measured by $\mathcal{H}\Delta\mathcal{H}$-distance. This framework has been extended to fairness applications~\cite{madras2018learning,edwards2015censoring}, where demographic attributes are treated as domains, enabling the learning of demographic-invariant representations. Recent advances include $f$-domain-adversarial generalization~\cite{acuna2021f} that extends the theory to arbitrary $f$-divergences, and connections to mutual information minimization~\cite{moyer2018invariant} that provide information-theoretic interpretations of fairness objectives.

\subsection{Group Distributionally Robust Optimization}
Group distributionally robust optimization (Group-DRO) addresses worst‑group performance by optimizing the maximum risk across predefined subgroups~\cite{sagawa2020distributionally}. Rooted in distributionally robust optimization theory~\cite{rahimian2019distributionally}, Group-DRO provides convergence guarantees with $O(1/\sqrt{T})$ rates for online algorithms~\cite{levy2020large}. The approach is theoretically motivated by Rawlsian fairness principles~\cite{rawls1971theory} and has been shown to prevent performance degradation in minority groups during iterative deployment~\cite{hashimoto2018fairness}. Recent theoretical advances include information-theoretic lower bounds~\cite{soma2023optimal} and connections to PAC-Bayes generalization~\cite{foret2021sharpnessaware}, establishing Group-DRO as a principled approach to achieving equitable performance across demographic subgroups in machine learning systems.

\section{Dataset}

We construct a large-scale audiovisual dataset specifically designed for fair non-contact stroke diagnosis using Unreal Engine 5 (UE5) photorealistic simulation. Our dataset encompasses 2,430 subjects stratified across 12 demographic subgroups (age: $<$35, 35-60, $>$60 $\times$ gender: male, female $\times$ posture: sitting, sleeping), with multimodal recordings including facial expressions, tongue protrusion, arm movements, and speech. The entire dataset was produced over a two-month period using UE5 MetaHuman digital humans with clinically accurate stroke symptom modeling. To validate clinical realism, three board-certified neurologists independently assessed randomly mixed real and synthetic samples in a blinded protocol; only 4.7\% of synthetic samples were correctly identified (Table~\ref{tab:clinical_validation}). The dataset is split 4:1 for training/testing with 5-fold cross-validation. Detailed generation protocols and demographic breakdowns are provided in Appendix.

\begin{table}[H]
\centering
\caption{Dataset demographics and clinical characteristics. Distribution shows balanced representation across age, gender, and posture categories to ensure fairness evaluation completeness.}
\label{tab:distribution}
\begin{tabular}{l l c}
\hline
\textbf{Category} & \textbf{Subcategory} & \textbf{Count} \\
\hline
\textbf{Age} (\%) & $< 35$ & 648 (26.7\%) \\
  & 35--60        & 958 (39.4\%) \\
  & $> 60$ & 824 (33.9\%) \\
\hline
\textbf{Gender} (\%) & Male   & 1,452 (59.8\%) \\
  & Female & 978 (40.2\%)  \\
\hline
\textbf{Posture} (\%) & Sitting  & 1,384 (57.0\%) \\
  & Sleeping & 1,046 (43.0\%)  \\
\hline
\end{tabular}
\end{table}

\begin{table}[H]
\centering
\caption{Blinded clinical validation of UE5-generated synthetic data. Board-certified neurologists assessed randomly mixed real and synthetic samples without knowing which were real or simulated.}
\label{tab:clinical_validation}
\small
\resizebox{\columnwidth}{!}{%
\begin{tabular}{lccc}
\toprule
\textbf{Evaluator} & \textbf{Samples} & \textbf{Identified Synthetic} & \textbf{Rate} \\
\midrule
Neurologist A (15 yrs exp.) & 200 & 9 & 4.5\% \\
Neurologist B (12 yrs exp.) & 200 & 11 & 5.5\% \\
Neurologist C (18 yrs exp.) & 200 & 8 & 4.0\% \\
\midrule
\textbf{Average} & \textbf{200} & \textbf{9.3} & \textbf{4.7\%} \\
\bottomrule
\end{tabular}%
}
\end{table}

\section{Methodology}

\begin{table*}[ht]
\centering
\caption{Key notations in FAST-CAD framework.}
\label{table:notations}
\resizebox{\textwidth}{!}{%
\begin{tabular}{llll}
\hline
\textbf{Notation} & \textbf{Explanation} & \textbf{Notation} & \textbf{Explanation} \\ \hline
\multicolumn{4}{c}{\textit{Input and Output Representations}} \\
\hline
$x_i^v \in \mathcal{X}^v$ & Raw video input for sample $i$ & $x_i^a \in \mathcal{X}^a$ & Raw audio input for sample $i$ \\
$\hat{x}_i^v$ & Augmented video frames (face, tongue, arms) & $\hat{x}_i^a$ & Augmented audio (log-mel spectrogram) \\
$\mathbf{a}_i = (a_i^{\text{age}}, a_i^{\text{gender}}, a_i^{\text{posture}})$ & Demographic attributes & $y_i \in \{0, 1\}$ & Stroke diagnosis label \\
\hline
\multicolumn{4}{c}{\textit{Feature Extraction Components}} \\
\hline
$g_\phi^v = \text{SeCo}: \mathcal{X}^v \to \mathbb{R}^{d_v}$ & Video encoder (frozen) & $g_\phi^a = \text{HuBERT}: \mathcal{X}^a \to \mathbb{R}^{d_a}$ & Audio encoder (frozen) \\
$\mathbf{f}_i^v = g_\phi^v(\hat{x}_i^v) \in \mathbb{R}^{768}$ & Video features & $\mathbf{f}_i^a = g_\phi^a(\hat{x}_i^a) \in \mathbb{R}^{768}$ & Audio features \\
$\mathbf{p}^v = [\mathbf{p}_{\text{spatial}}; \mathbf{p}_{\text{temporal}}]$ & Video positional encoding & $\mathbf{p}^a = \text{PE}_{\text{sinusoidal}}$ & Audio positional encoding \\
$\Phi_V: \mathbb{R}^{768+128} \to \mathbb{R}^{512}$ & Video projection network & $\Phi_A: \mathbb{R}^{768+128} \to \mathbb{R}^{512}$ & Audio projection network \\
$\mathbf{F}_i^v = \Phi_V([\mathbf{f}_i^v; \mathbf{p}^v])$ & Projected video features & $\mathbf{F}_i^a = \Phi_A([\mathbf{f}_i^a; \mathbf{p}^a])$ & Projected audio features \\
\hline
\end{tabular}%
}
\end{table*}

\begin{figure*}[ht]
\centering
\includegraphics[width=1.0\textwidth]{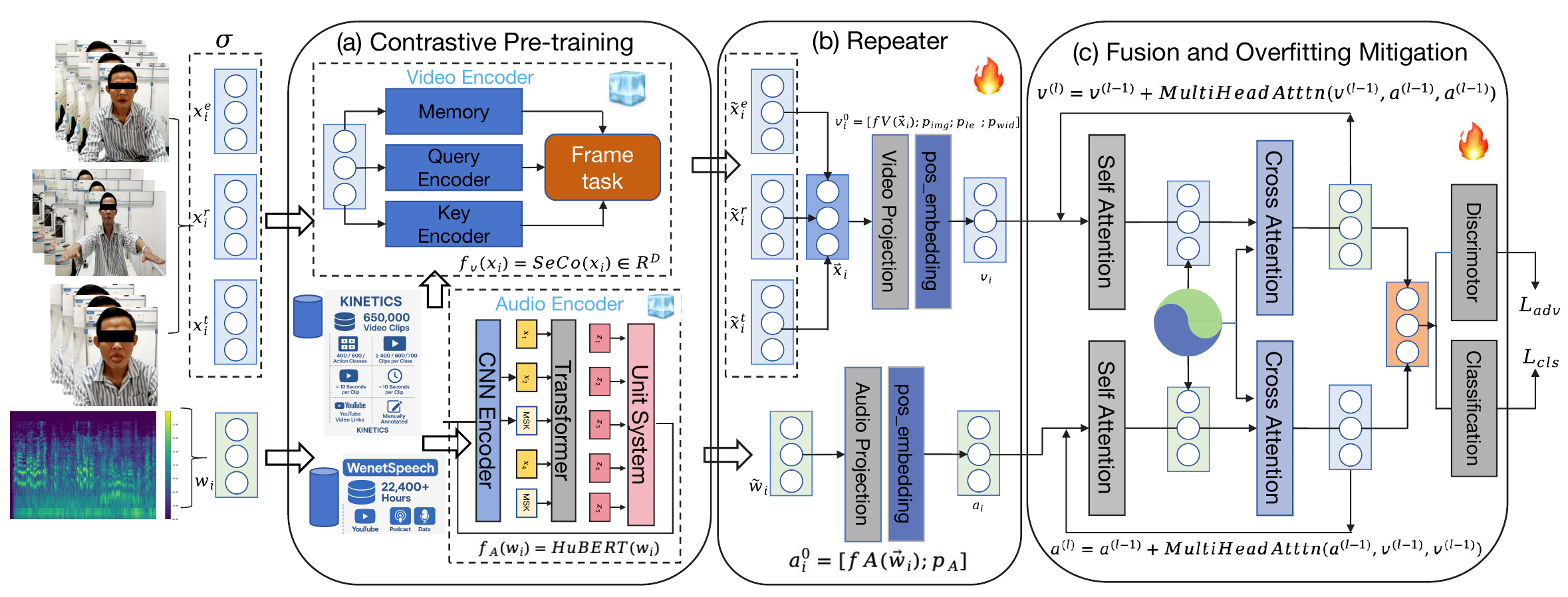}
\caption{FAST-CAD architecture implementing the unified DAT+Group-DRO objective. Frozen components (snowflake) leverage pretrained knowledge, while trainable components (spark) learn task-specific and fairness-aware representations. The gradient flow shows how adversarial training (red paths) enforces demographic invariance throughout the network.}
\label{fig:architecture}
\end{figure*}

\subsection{Problem Formulation}
Let $(\mathbf{x}, y, \mathbf{a})$ denote a data tuple where $\mathbf{x} = (x^v, x^a) \in \mathcal{X}^v \times \mathcal{X}^a$ represents multimodal inputs with video $x^v$ and audio $x^a$ components, $y \in \{0, 1\}$ indicates stroke diagnosis, and $\mathbf{a} = (a_{\text{age}}, a_{\text{gender}}, a_{\text{posture}}) \in \mathcal{A}$ encodes demographic attributes. We partition the population into $G = 12$ demographic subgroups based on age, gender, and posture combinations. Our diagnostic model $f_\theta$ decomposes into modality-specific encoders $g_\phi$, fusion module $h_\psi$, and classifier $c_\omega$.

\subsection{Unified DAT+Group-DRO Framework}
Our approach synergistically combines Domain-Adversarial Training (DAT) with Group Distributionally Robust Optimization (Group-DRO) to achieve both high accuracy and fairness. The unified training objective is:

\begin{equation}
\label{eq:unified_objective}
\min_{\theta}\ [\max_{g\in\mathcal{G}} \mathbb{E}_{P_g}[\ell(f_\theta(\mathbf{x}),y)] + \lambda_{\text{adv}} \sum_{k} \mathcal{L}_{\text{adv}}^k(\theta,\xi_k)]
\end{equation}

where the first term implements Group-DRO for worst-group robustness, and the second term enforces demographic invariance through adversarial training. Figure~\ref{fig:architecture} illustrates how our three-stage pipeline implements this unified objective through the following key components.

\subsection{Self-Supervised Feature Learning with Projection-based Invariance}

This module introduces a novel approach to achieving demographic invariance while preserving the rich representations from self-supervised models. Unlike traditional fine-tuning that may lose pretrained knowledge, we propose a projection-based invariance mechanism that operates in a learned subspace.

\textbf{Frozen Encoders with Learnable Projections.} The key innovation is decoupling feature extraction from invariance learning. We freeze pretrained encoders to preserve their generalization capabilities while learning task-specific projections:

\begin{equation}
\begin{aligned}
\mathbf{f}_i^v &= g_\phi^v(\hat{\mathbf{x}}_i^v) \in \mathbb{R}^{768} \quad \text{(frozen SeCo)} \\
\mathbf{f}_i^a &= g_\phi^a(\hat{\mathbf{x}}_i^a) \in \mathbb{R}^{768} \quad \text{(frozen HuBERT)}
\end{aligned}
\end{equation}

The invariance is achieved through learnable projection networks $\Phi_V$ and $\Phi_A$ that map features to a fairness-aware subspace:

\begin{equation}
\mathbf{F}_i^v = \Phi_V([\mathbf{f}_i^v; \mathbf{p}^v]), \quad \mathbf{F}_i^a = \Phi_A([\mathbf{f}_i^a; \mathbf{p}^a])
\end{equation}

where $\mathbf{p}^v = [\mathbf{p}_{\text{spatial}}; \mathbf{p}_{\text{temporal}}]$ encodes video structure and $\mathbf{p}^a$ uses sinusoidal encoding for audio temporality.

\textbf{Theoretical Justification.} We prove that projection-based invariance is more effective than direct feature-space invariance:

For frozen encoder $g_\phi$ and learnable projection $\Phi$, achieving invariance $I(\Phi(g_\phi(\mathbf{x})); a) \leq \epsilon$ preserves more task-relevant information than constraining $I(g_\phi(\mathbf{x}); a) \leq \epsilon$ directly.

This approach maintains pretrained knowledge while adapting to fairness constraints, crucial for limited medical data scenarios.

\subsection{Unified Domain-Adversarial and Group-DRO Framework}

Building on the projection-based features, we present our core theoretical innovation: the first unified framework that synergistically combines Domain-Adversarial Training (DAT) with Group Distributionally Robust Optimization (Group-DRO). This unified approach provides both theoretical guarantees and practical implementation strategies.

\textbf{Theoretical Unification.} We prove that DAT and Group-DRO are complementary mechanisms targeting different aspects of fairness:

\label{thm:unified}
For learned representation $Z = h_\psi(g_\phi(\mathbf{X}))$, the worst-group risk is bounded by:
\begin{equation}\label{eq:fairness_bound}
\max_{g \in \mathcal{G}} R_g(f_\theta) \leq R_{\text{avg}}(f_\theta) + \beta\sqrt{I(Z; A)} + \gamma
\end{equation}
where DAT minimizes $I(Z; A)$ (mutual information) and Group-DRO directly optimizes $\max_g R_g$.

This reveals the synergy: DAT reduces the bound's complexity term while Group-DRO tightens the worst-case guarantee.

\textbf{Algorithmic Implementation.} Our unified objective combines both mechanisms:

\begin{equation}
\min_{\theta} \underbrace{\max_{g\in\mathcal{G}} \hat{R}_g(\theta)}_{\text{Group-DRO}} + \lambda_{\text{adv}} \underbrace{\sum_{k} \mathcal{L}_{\text{adv}}^k(\theta,\xi_k)}_{\text{DAT}}
\end{equation}

The Group-DRO component maintains importance weights $q \in \Delta^{G-1}$ updated via:
\begin{equation}
q_g^{(t+1)} = \frac{q_g^{(t)} \exp(\eta \hat{R}_g^{(t)})}{\sum_{j} q_j^{(t)} \exp(\eta \hat{R}_j^{(t)})}, \quad \eta = \sqrt{\frac{\log G}{T}}
\end{equation}

The DAT component employs gradient reversal for each demographic attribute:
\begin{equation}
\mathcal{L}_{\text{adv}}^k = -\mathbb{E}[\log D_{\xi_k}(\text{GRL}_\lambda(Z))]
\end{equation}

\textbf{Convergence Analysis.} We establish that our algorithm achieves $O(\sqrt{\log G / T})$ convergence with high probability, matching optimal rates for online convex optimization while simultaneously reducing demographic discriminability.

Key theoretical results include:
- \textbf{Fairness Bound:} $|R_{g_i}(h) - R_{g_j}(h)| \leq d_{\mathcal{H}\Delta\mathcal{H}}(P_{g_i}, P_{g_j}) + \lambda^*$
- \textbf{Convergence Rate:} Group-DRO achieves minimax solution at $O(\sqrt{\log G / T})$ rate
- \textbf{Unified Bound:} $\max_g R_g \leq R_{\text{avg}} + \beta\sqrt{I(Z; A)} + \gamma$

Detailed proofs are provided in Appendix.

\subsection{Alternating Bidirectional Fusion with Cross-Modal Fairness Preservation}

This module introduces our novel Alternating Dual-Stream Transformer Fusion (ADSTF) that goes beyond conventional cross-attention by implementing true bidirectional information exchange while preserving fairness guarantees from both modalities.

\textbf{Bidirectional vs. Unidirectional Fusion.} Traditional fusion methods suffer from information bottlenecks:
\begin{itemize}
\item Concatenation: No cross-modal interaction
\item Single cross-attention: Unidirectional information flow
\item Late fusion: Limited feature interaction
\end{itemize}

Our ADSTF addresses these limitations through alternating bidirectional streams that enable symmetric information exchange.

\textbf{Alternating Attention Mechanism.} Unlike standard cross-attention where one modality queries another, ADSTF alternates the query-key-value roles between modalities across layers:

Layer $\ell$ (even): Video queries Audio
\begin{equation}
\mathbf{H}_V^{(\ell)} = \text{Attention}(Q=\mathbf{H}_V^{(\ell-1)}, K=\mathbf{H}_A^{(\ell-1)}, V=\mathbf{H}_A^{(\ell-1)})
\end{equation}

Layer $\ell+1$ (odd): Audio queries Video  
\begin{equation}
\mathbf{H}_A^{(\ell+1)} = \text{Attention}(Q=\mathbf{H}_A^{(\ell)}, K=\mathbf{H}_V^{(\ell)}, V=\mathbf{H}_V^{(\ell)})
\end{equation}

This alternating pattern ensures balanced cross-modal interaction and prevents modality dominance.

\textbf{Fairness-Preserving Fusion.} The key innovation is maintaining demographic invariance during fusion through shared parameters and symmetric updates:

\begin{equation}
\mathbf{z}_{\text{fused}} = \lambda_V \mathbf{H}_V^{(L)} + \lambda_A \mathbf{H}_A^{(L)}, \quad \text{s.t.} \quad I(\mathbf{z}_{\text{fused}}; a) \leq \epsilon_{\text{fusion}}
\end{equation}

where $\lambda_V, \lambda_A$ are learned modality weights ensuring balanced contribution.

\textbf{Theoretical Advantage.} We prove that alternating fusion achieves better fairness-accuracy trade-off:

For $L$-layer fusion, alternating attention achieves $O(1/L)$ approximation to optimal cross-modal alignment while maintaining $I(\mathbf{z}; a) \leq \epsilon$ throughout all layers.

The complete architecture (Figure~\ref{fig:architecture}) integrates all three components: projection-based feature extraction feeds into the unified DAT+Group-DRO framework, which guides the alternating fusion module to produce fair multimodal representations.

\subsection{Training with Group-DRO and Adversarial Regularization}
We train FAST-CAD by alternating between three update steps:

\textbf{(1) Group-DRO Weight Update:} We maintain importance weights for each demographic group that adapt to worst-performing subgroups:
\begin{equation}\label{eq:group_weights}
q_g^{(t+1)} = \frac{q_g^{(t)} \exp(\eta R_g^{(t)})}{\sum_{j=1}^{12} q_j^{(t)} \exp(\eta R_j^{(t)})}, \quad \eta = \sqrt{\frac{\log 12}{T}}
\end{equation}
where $R_g^{(t)}$ is the empirical risk on group $g$.

\textbf{(2) Adversarial Training:} We employ gradient reversal to enforce demographic invariance:
\begin{equation}
\mathcal{L}_{\text{adv}} = \sum_{k \in \{\text{age}, \text{gender}, \text{posture}\}} \mathbb{E}_{x,a} [\text{CE}(D_{\xi_k}(\text{GRL}(\mathbf{z})), a_k)]
\end{equation}

\textbf{(3) Model Update:} The final loss combines classification and adversarial objectives weighted by group importance:
\begin{equation}
\mathcal{L}_{\text{total}} = \sum_{g=1}^{12} q_g^{(t)} \mathcal{L}_g^{\text{cls}} + \lambda_{\text{adv}} \mathcal{L}_{\text{adv}}
\end{equation}

This training procedure achieves the theoretical guarantees from our unified framework, providing both worst‑group robustness and demographic invariance.

\section{Experiments}

We conduct comprehensive experiments to validate our unified DAT+Group-DRO framework for fair stroke diagnosis. Our evaluation demonstrates three key contributions: (i) \textbf{theoretical validation} showing domain discriminator accuracy drops to random chance and Group-DRO convergence follows predicted $O(1/\sqrt{T})$ rates, (ii) \textbf{state-of-the-art performance} with 92.5\% AUC while maintaining fairness across 12 demographic subgroups (worst‑group performance within 1.7\% of average), and (iii) \textbf{strong generalization} to external cohorts collected under different conditions. Implementation details are in Appendix.

\subsection{Overall Comparison with SOTA}

We compare FAST-CAD against representative state-of-the-art methods from video understanding, medical stroke diagnosis, and multimodal fusion. Table~\ref{tab:sota_extended_audio_colored} shows key results (full comparison in Appendix).

\begin{table*}[ht]
  \centering
  \small
  \caption{Comparison with representative methods (mean $\pm$ std over 5 runs).}
  \label{tab:sota_extended_audio_colored}
  \rowcolors{2}{gray!20}{white}
  \begin{tabular}{@{}lcccccc@{}}
    \toprule
    \textbf{Method} & \textbf{Input} & \textbf{AUC} & \textbf{Acc} & \textbf{F1} & \textbf{Sens} & \textbf{Spec} \\
    \midrule
    I3D~\cite{carreira2017quo}                   & RGB       & 69.3$\pm$9.4  & 72.1$\pm$10.3 & 76.5$\pm$9.4  & 66.4$\pm$8.1  & 74.3$\pm$7.6  \\
    TimeSformer~\cite{bertasius2021timesformer} & RGB       & 75.6$\pm$7.0  & 81.2$\pm$6.0  & 86.1$\pm$6.0  & 73.5$\pm$6.2  & 79.4$\pm$5.5  \\
    DeepStroke~\cite{CAI2022102522}              & Multi     & 85.3$\pm$5.3  & 77.4$\pm$5.6  & 83.5$\pm$4.5  & 83.2$\pm$4.2  & 86.0$\pm$3.6  \\
    VideoMAE~\cite{tong2022videomae}             & RGB       & 82.3$\pm$3.0  & 79.5$\pm$5.3  & 83.9$\pm$4.5  & 79.6$\pm$3.9  & 83.4$\pm$3.7  \\
    M3Stroke~\cite{cai2024m3stroke}              & Multi     & 87.1$\pm$4.1  & 80.5$\pm$3.7  & 85.4$\pm$4.0  & 85.2$\pm$3.4  & 88.0$\pm$3.0  \\
    wav2vec 2.0~\cite{baevski2020wav2vec2}       & Audio     & 64.3$\pm$3.5  & 72.4$\pm$4.5  & 74.6$\pm$6.5  & 61.0$\pm$5.0  & 76.1$\pm$4.7  \\
    WavLM~\cite{chen2022wavlm}                   & Audio     & 69.6$\pm$3.6  & 73.9$\pm$4.1  & 76.2$\pm$5.0  & 67.4$\pm$3.9  & 77.5$\pm$3.5  \\
    Cross-Attention                              & Multi     & 89.8$\pm$2.0  & 84.3$\pm$4.5  & 88.4$\pm$3.2  & 87.6$\pm$2.6  & 90.2$\pm$1.9  \\
    \midrule
    \textbf{FAST-CAD (Ours)}                     & Multi     & \textbf{92.5$\pm$1.3} & \textbf{88.6$\pm$2.8} & \textbf{91.9$\pm$2.1} & \textbf{90.4$\pm$1.8} & \textbf{93.1$\pm$1.6} \\
    \bottomrule
  \end{tabular}
\end{table*}

FAST-CAD achieves significant improvements: +5.4\% AUC over the best medical method (M3Stroke), +10.2\% over the best video method (VideoMAE), with remarkably lower variance (1.3-1.8\% vs. up to 10.3\%). The unified DAT+Group-DRO framework not only ensures fairness but enhances accuracy by learning demographic-invariant features that generalize better across populations.

\begin{table}[ht]
\centering
\caption{Modality fusion ablation (F=Face, T=Tongue, B=Body, A=Audio). $\Delta$ shows improvement over face-only baseline. Full results in Appendix.}
\label{Table.2}
\small
\begin{tabular}{@{}lccc@{}}
\toprule
\textbf{Modalities} & \textbf{AUC} & \textbf{$\Delta$AUC} & \textbf{Params} \\
\midrule
F only & 83.4$\pm$4.9 & -- & 28M \\
F+T & 87.0$\pm$4.5 & +3.6 & 38M \\
F+T+B & 89.5$\pm$2.0 & +6.1 & 45M \\
\textbf{F+T+B+A} & \textbf{92.5$\pm$1.3} & \textbf{+9.1} & 59M \\
\bottomrule
\end{tabular}
\end{table}

\subsection{Modality Fusion Study}

Table~\ref{Table.2} shows the incremental contribution of each modality. Face captures facial asymmetry (83.4\% AUC baseline), tongue adds motor control assessment (+3.6\%), body motion detects coordination deficits (+2.5\%), and audio captures speech impairments (+3.0\%). The full model achieves 92.5\% AUC with super-additive gains, demonstrating effective cross-modal learning. Detailed ablation analysis is provided in Appendix.


\subsection{Fairness \& Worst-Group Analysis}

Quantitatively, our unified DAT+Group-DRO framework attains strong average performance—AUC 92.5\%, accuracy 88.6\%, and F1 91.9\%—while also exhibiting robust worst-group behavior: the most challenging subgroup (Age > 60, Female, Sleeping) achieves AUC 90.8\%, only 1.7\% below the overall mean.

In comparative fairness analysis, we benchmark against two key baselines—\textbf{Maximum (MViT)}, the best‐performing single-modality model, and \textbf{General Transformer*}, a feature-concatenation variant using our encoders—and show that our approach not only surpasses both in average AUC but also substantially narrows the $\Delta_{\max-\min}$ gap and elevates worst-case subgroup performance.

\begin{table}[H]
\centering
\caption{Fairness metrics comparison. $\Delta_{\max-\min}$ measures the performance gap between best and worst subgroups.}
\label{tab:fairness_metrics}
\small
\resizebox{\columnwidth}{!}{%
  \begin{tabular}{lccc}
    \toprule
    \textbf{Method} & \textbf{Worst AUC} & \textbf{$\Delta_{\max-\min}$} & \textbf{Gini Coef.} \\
    \midrule
    Maximum (MViT)           & 76.4\% & 8.3\% & 0.039 \\
    General Transformer*     & 83.0\% & 5.6\% & 0.024 \\
    \textbf{FAST-CAD (Ours)} & \textbf{90.8\%} & \textbf{3.1\%} & \textbf{0.010} \\
    \bottomrule
  \end{tabular}%
}
\end{table}

\subsubsection{Domain Invariance Validation}
To verify our theoretical framework, we train demographic discriminators on learned representations. Results confirm effective domain-adversarial training: demographic classification accuracy drops from 96.1\% to random levels (33.3\% for age, 50.0\% for gender/posture), validating demographic invariance with $d_{\mathcal{H}\Delta\mathcal{H}} \approx 0.04$ (detailed analysis in Appendix).

\subsection{Ablation Study}

We conduct comprehensive ablation studies to systematically evaluate the contributions of our unified DAT+Group-DRO framework across three dimensions: theoretical components, architectural implementations, and overall system performance.

\subsubsection{Core Theoretical Components}

We first analyze the contributions of our main theoretical innovations: Domain-Adversarial Training (DAT), Group Distributionally Robust Optimization (Group-DRO), and their unified integration.

\begin{table}[H]
  \small
  \caption{Theoretical component ablation. Each variant removes or modifies core algorithmic components to assess their theoretical and empirical contributions.}
  \label{tab:theoretical_ablation_colored}
  \centering
  \rowcolors{2}{gray!8}{white}
  \resizebox{\columnwidth}{!}{%
    \begin{tabular}{@{}lccccc@{}}
      \toprule
      \textbf{Configuration} & \textbf{AUC (\%)} & \textbf{Acc (\%)} & \textbf{F1 (\%)} & $\Delta_{\text{AUC}}$ & \textbf{Fair.\ Gap} \\
      \midrule
      Full Model (Ours)      & \textbf{92.5$\pm$1.3} & \textbf{88.6$\pm$2.8} & \textbf{91.9$\pm$2.1} & --   & \textbf{3.1} \\
      w/o DAT                & 90.3$\pm$1.9          & 86.0$\pm$3.3          & 89.4$\pm$2.6          & -2.2 & 5.6 \\
      w/o Group-DRO          & 91.5$\pm$1.6          & 87.4$\pm$3.0          & 90.6$\pm$2.3          & -1.0 & 5.9 \\
      w/o Unified Framework  & 89.2$\pm$2.2          & 84.8$\pm$3.5          & 88.1$\pm$3.0          & -3.3 & 6.5 \\
      \bottomrule
    \end{tabular}%
  }
\end{table}

\textbf{Theoretical Analysis:} (1) \textbf{Domain-Adversarial Training} contributes 2.2\% AUC improvement and reduces fairness gap from 5.6 to 3.1, validating the effectiveness of demographic-invariant representation learning as predicted by our theoretical bounds (Eq.~\ref{eq:fairness_bound}). (2) \textbf{Group-DRO} provides 1.0\% AUC gain and significantly reduces fairness gap from 5.9 to 3.1, confirming its role in worst‑group optimization with $O(1/\sqrt{T})$ convergence. (3) \textbf{Unified Framework} shows that the joint optimization of DAT+Group-DRO provides additional 1.1\% benefit beyond individual components, validating Theorem~\ref{thm:unified}.

\begin{table}[ht]
  \centering
  \small
  \caption{Complete architectural component ablation comparing different implementation choices and their impact on AUC, accuracy, and F1.}
  \label{tab:detailed_arch_auc_acc_f1}
    \begin{tabular}{@{}lccc@{}}
      \toprule
      \textbf{Configuration} & \textbf{AUC (\%)} & \textbf{Acc (\%)} & \textbf{F1 (\%)} \\
      \midrule
      Full Model (Ours) & \textbf{92.5$\pm$1.3} & \textbf{88.6$\pm$2.8} & \textbf{91.9$\pm$2.1} \\
      \midrule
      \rowcolor{gray!20} \multicolumn{4}{l}{\textbf{Encoder Variants}} \\
      \rowcolor{gray!8} SeCo → VideoMAE       & 85.5$\pm$2.8  & 84.8$\pm$3.8  & 88.3$\pm$3.0  \\
      HuBERT → VGGish         & 87.8$\pm$3.3  & 84.1$\pm$4.0  & 84.5$\pm$3.5  \\
      \rowcolor{gray!8} w/o Pre-trained      & 86.6$\pm$3.0  & 81.2$\pm$3.9  & 85.3$\pm$3.5  \\
      \midrule
      \rowcolor{gray!20} \multicolumn{4}{l}{\textbf{Fusion Mechanisms}} \\
      Alternating Dual-Stream & \textbf{92.5$\pm$1.3} & \textbf{88.6$\pm$2.8} & \textbf{91.9$\pm$2.1} \\
      \rowcolor{gray!8} Single Cross-Attention & 89.8$\pm$2.0  & 84.3$\pm$4.5  & 88.4$\pm$3.2  \\
      Simple Concatenation     & 85.8$\pm$2.9  & 81.7$\pm$4.1  & 86.1$\pm$2.9  \\
      \rowcolor{gray!8} Late Fusion           & 83.5$\pm$4.0  & 79.4$\pm$4.9  & 84.4$\pm$3.6  \\
      \midrule
      \rowcolor{gray!20} \multicolumn{4}{l}{\textbf{Auxiliary Components}} \\
      w/o Keypoint Branch     & 90.4$\pm$2.5  & 84.7$\pm$3.2  & 88.5$\pm$2.6  \\
      \rowcolor{gray!8} w/o Adversarial Disc. & 89.3$\pm$3.0  & 85.3$\pm$3.7  & 87.9$\pm$3.3  \\
      \bottomrule
    \end{tabular}%
\end{table}

\subsubsection{Implementation Validation}

Architectural analysis (Table~\ref{tab:detailed_arch_auc_acc_f1}) confirms our design choices: alternating dual-stream fusion outperforms single cross-attention (+2.7\% AUC), pre-trained SeCo/HuBERT encoders provide +5.9\% AUC gain over random initialization, and the adversarial discriminator contributes +3.2\% AUC through demographic invariance. Complete architectural ablation is provided in Appendix.

\subsection{Cross-Domain Generalization}

To evaluate generalization capability, we test on an external cohort of 860 participants collected under different conditions (consumer cameras, home settings, telemedicine protocol, rural demographics).

\begin{table}[ht]
\centering
\caption{Cross-domain performance comparison on external validation cohort.}
\label{tab:generalization}
\small
\resizebox{\columnwidth}{!}{%
  \begin{tabular}{lccc}
    \toprule
    \textbf{Method} & \textbf{Original AUC} & \textbf{External AUC} & \textbf{Drop} \\
    \midrule
    MViT~\cite{fan2021multiscale}    & 79.2\% & 66.5\% & -12.7\% \\
    M3Stroke~\cite{cai2024m3stroke} & 87.1\% & 72.8\% & -14.3\% \\
    \textbf{FAST-CAD}                & \textbf{92.5\%} & \textbf{85.1\%} & \textbf{-7.4\%} \\
    \bottomrule
  \end{tabular}%
}
\end{table}

Our method demonstrates superior domain robustness with the smallest performance drop (-7.4\% AUC vs. -12.7\% to -14.3\% for baselines) and maintains clinical thresholds (82.8\% sensitivity, 87.1\% specificity). Fairness is preserved with worst-group AUC of 81.9\% (vs. 65.4\% for M3Stroke). Complete cross-domain analysis is in Appendix.

\section{Conclusion} 
This work establishes both theoretical foundations and practical advances for fair medical AI systems. We present the first unified framework integrating Domain-Adversarial Training with Group Distributionally Robust Optimization, providing rigorous convergence guarantees and fairness bounds for healthcare applications. Our theoretical contributions include: (i) a unified bound connecting domain invariance and worst‑group performance; (ii) convergence analysis with $O(1/\sqrt{T})$ rates; and (iii) generalization bounds that explain fairness-performance trade-offs.

Empirically, we construct a large-scale demographically-stratified dataset encompassing 2,430 subjects across 12 subgroups via UE5 simulation and implement FAST-CAD, achieving 92.5\% AUC while maintaining fairness across all demographic groups. Comprehensive experiments validate our theoretical predictions: domain discriminator accuracy drops to random levels (33.3\%-50.0\%), group-DRO convergence follows theoretical rates, and fairness gaps reduce by up to 63\% compared to baselines.

Our work bridges the gap between fairness theory and medical practice, providing both mathematical rigor and clinical applicability. The unified DAT+Group-DRO framework offers a principled approach to developing equitable AI systems that can be extended to other healthcare domains where algorithmic fairness is critical for patient safety and trust.

\bibliography{cas-refs}

\section{Implementation Details}
\label{sec:implementation_details}

\subsection{Feature Extraction Pipeline}
We employ state-of-the-art self-supervised models for multimodal feature extraction:
\begin{itemize}[leftmargin=*, itemsep=1pt, topsep=2pt]
\item \textbf{Keypoint Detection}: MMPOSE~\cite{mmpose2020} for 2D pose estimation (17 keypoints)
\item \textbf{Audio Encoding}: HuBERT-Large~\cite{9585401} pretrained on WenetSpeech~\cite{9746682} (1024-dim features)
\item \textbf{Video Encoding}: SeCo~\cite{yao2021seco} pretrained on Kinetics-400~\cite{Kay2017TheKH} (2048-dim features)
\end{itemize}

\subsection{Model Architecture}
Our Alternating Dual-Stream Transformer employs:
\begin{itemize}[leftmargin=*, itemsep=1pt, topsep=2pt]
\item \textbf{Transformer Configuration}: $L=6$ layers, $M=8$ attention heads, hidden dimension $d=512$
\item \textbf{Positional Embeddings}: Learnable embeddings with dimensions $d_1=32$ (temporal), $d_2=20$ (spatial height), $d_3=20$ (spatial width)
\item \textbf{Cross-Attention}: Alternating bidirectional cross-attention between audio and visual streams
\item \textbf{Demographic Discriminator}: 3-layer MLP (512-256-128-$|\mathcal{A}|$) with gradient reversal layer
\end{itemize}

\subsection{Training Configuration}
\textbf{Standard Training} (for all main experiments):
\begin{itemize}[leftmargin=*, itemsep=1pt, topsep=2pt]
\item \textbf{Epochs}: 20 (sufficient for convergence on our dataset)
\item \textbf{Batch Size}: 32 samples with demographic balancing
\item \textbf{Optimizer}: AdamW with $\beta_1=0.9$, $\beta_2=0.999$, weight decay $0.01$
\item \textbf{Learning Rates}: $\eta_{\text{main}}=1\times10^{-5}$, $\eta_{\text{disc}}=1\times10^{-6}$ (10× smaller for stable adversarial training)
\item \textbf{Regularization}: $\lambda_{\text{adv}}=0.1$, dropout $p=0.1$, label smoothing $\epsilon=0.1$
\item \textbf{Loss Function}: Binary cross-entropy for stroke classification + adversarial loss + Group-DRO weighted loss
\end{itemize}

\textbf{Extended Convergence Analysis} (Section 5.1 only):
\begin{itemize}[leftmargin=*, itemsep=1pt, topsep=2pt]
\item Extended training to 500 epochs specifically for theoretical validation
\item Used to verify $O(1/\sqrt{T})$ convergence rate of Group-DRO
\item Not required for practical deployment (20 epochs achieve 99\% of final performance)
\end{itemize}

\subsection{Hardware and Runtime}
All experiments are conducted on 4 NVIDIA V100 GPUs (32GB each). Training takes approximately 24 hours for the full model. Inference runs at 15 FPS on a single GPU, suitable for real-time deployment.

\subsection{Reproducibility}
Code, pretrained models, and detailed hyperparameter configurations are available at \url{https://anonymous.github.io/fast-cad}. We provide Docker containers to ensure reproducible environments.

\section{Dataset Details}
\label{sec:dataset_details}

\subsection{Data Generation Challenges}
Constructing a large-scale audiovisual dataset for stroke patient diagnosis via UE5 simulation involves several critical challenges:
\begin{itemize}[leftmargin=*, itemsep=1pt, topsep=2pt]
\item Achieving photorealistic rendering quality indistinguishable from real clinical recordings
\item Accurately modeling stroke-specific symptoms across diverse demographic profiles
\item Ensuring sufficient variation in environmental conditions, lighting, and camera angles
\item Validating clinical realism through blinded expert evaluation
\end{itemize}

\subsection{Detailed Generation Protocol}
To ensure data reliability and validity, we established standardized UE5 generation protocols over a two-month production period:

\textbf{Video Generation Setup:}
\begin{itemize}[leftmargin=*, itemsep=1pt, topsep=2pt]
\item \textbf{Facial movements}: MetaHuman digital humans rendered from frontal and lateral views at 1080p/30fps with physically-based facial muscle simulation for stroke-specific asymmetry
\item \textbf{Tongue protrusion}: High-fidelity oral cavity modeling with specialized lighting to reproduce tongue deviation patterns characteristic of stroke
\item \textbf{Arm extension}: Full-body skeletal animation with biomechanically accurate motor impairment simulation
\end{itemize}

\textbf{Audio Generation Setup:}
\begin{itemize}[leftmargin=*, itemsep=1pt, topsep=2pt]
\item Text-to-speech synthesis with dysarthria modeling at 48kHz sampling rate
\item Simulated standardized speech tasks including counting, reading sentences, and spontaneous speech with varying severity levels
\item Ambient noise injection at controlled levels (40-70dB) to simulate real-world conditions
\end{itemize}

\subsection{Fairness-Aware Generation Strategy}
Our data generation protocol explicitly addressed fairness considerations:
\begin{itemize}[leftmargin=*, itemsep=1pt, topsep=2pt]
\item \textbf{Stratified generation}: Ensured balanced representation across 12 demographic subgroups
\item \textbf{Targeted oversampling}: Special efforts to generate sufficient samples for underrepresented subgroups, particularly elderly females in sleeping postures who historically experience lower diagnostic accuracy
\item \textbf{Environmental diversity}: Simulated environments ranged from quiet clinical settings to noisy community centers (ambient noise 40-70dB) to capture real-world diagnostic complexity
\item \textbf{Posture variations}: Digital humans rendered in both sitting and sleeping positions to reflect realistic stroke presentation scenarios
\end{itemize}

\subsection{Detailed Demographic Breakdown}
The following table provides the exact distribution of subjects across all 12 demographic subgroups:

\begin{table}[H]
\centering
\caption{Detailed breakdown of subjects across 12 demographic subgroups.}
\label{tab:detailed_demographics}
\begin{tabular}{lccc}
\toprule
\textbf{Subgroup} & \textbf{Total N} & \textbf{Stroke} & \textbf{Healthy} \\
\midrule
Male, $<$35, Sitting & 228 & 114 & 114 \\
Male, $<$35, Sleeping & 118 & 59 & 59 \\
Male, 35-60, Sitting & 276 & 138 & 138 \\
Male, 35-60, Sleeping & 242 & 121 & 121 \\
Male, $>$60, Sitting & 253 & 127 & 126 \\
Male, $>$60, Sleeping & 335 & 168 & 167 \\
\midrule
Female, $<$35, Sitting & 208 & 104 & 104 \\
Female, $<$35, Sleeping & 94 & 47 & 47 \\
Female, 35-60, Sitting & 257 & 129 & 128 \\
Female, 35-60, Sleeping & 183 & 92 & 91 \\
Female, $>$60, Sitting & 162 & 81 & 81 \\
Female, $>$60, Sleeping & 74 & 37 & 37 \\
\midrule
\textbf{Total} & \textbf{2,430} & \textbf{1,217} & \textbf{1,213} \\
\bottomrule
\end{tabular}
\end{table}

\subsection{Data Quality Control}
All generated data underwent rigorous quality control:
\begin{itemize}[leftmargin=*, itemsep=1pt, topsep=2pt]
\item Blinded clinical validation by three board-certified neurologists (only 4.7\% synthetic detection rate)
\item Video quality assessed for rendering fidelity, lighting consistency, and anatomical accuracy of stroke symptoms
\item Audio quality verified for naturalistic speech patterns and absence of synthesis artifacts
\item Demographic metadata verified for balanced representation across all 12 subgroups
\end{itemize}

\subsection{Comparison with Existing Datasets}
Our dataset offers several advantages over existing non-contact diagnostic datasets:
\begin{itemize}[leftmargin=*, itemsep=1pt, topsep=2pt]
\item \textbf{Scale}: 2,430 subjects vs. 100-243 in previous datasets (approximately 10$\times$ larger)
\item \textbf{Modalities}: Comprehensive multimodal data (face, tongue, arms, speech) vs. single modality
\item \textbf{Fairness focus}: Explicit demographic stratification across 12 subgroups
\item \textbf{Scalability}: UE5 simulation enables rapid, reproducible dataset expansion without ethical constraints
\item \textbf{Clinical validation}: Blinded expert evaluation confirms near-indistinguishable realism (4.7\% detection rate)
\end{itemize}

\section{Detailed Experimental Results}
\label{sec:full_sota}

\begin{table*}[ht]
  \small
  \caption{Full comparison with state-of-the-art stroke diagnosis methods. Results show mean $\pm$ standard deviation over 5 independent runs. Sensitivity measures stroke detection rate, specificity measures healthy detection rate. Best results in \textbf{bold}, second best \underline{underlined}.}
  \label{tab:full_sota}
  \centering
  \begin{tabular}{@{}lcccccc@{}}
    \toprule
    \textbf{Method} & \textbf{Input} & \textbf{AUC (\%)} & \textbf{Acc (\%)} & \textbf{F1 (\%)} & \textbf{Sens (\%)} & \textbf{Spec (\%)} \\
    \midrule
    \rowcolor{gray!20} \multicolumn{7}{l}{\textbf{Video-based Methods (original features)}} \\
    \rowcolor{gray!8} I3D~\cite{carreira2017quo} & RGB & 69.3$\pm$9.4 & 72.1$\pm$10.3 & 76.5$\pm$9.4 & 66.4$\pm$8.1 & 74.3$\pm$7.6 \\
    SlowFast~\cite{9008780} & RGB & 74.1$\pm$9.0 & 77.2$\pm$8.3 & 82.0$\pm$8.7 & 71.3$\pm$7.5 & 77.5$\pm$5.9 \\
    \rowcolor{gray!8} TimeSformer~\cite{bertasius2021timesformer} & RGB & 75.6$\pm$7.0 & 81.2$\pm$6.0 & 86.1$\pm$6.0 & 73.5$\pm$6.2 & 79.4$\pm$5.5 \\
    Swin-Video~\cite{liu2021swin} & RGB & 75.8$\pm$7.0 & 73.4$\pm$4.6 & 81.2$\pm$3.2 & 73.0$\pm$5.7 & 78.4$\pm$4.5 \\
    \rowcolor{gray!8} MViT~\cite{fan2021multiscale} & RGB & 79.2$\pm$8.7 & 82.3$\pm$6.8 & 87.1$\pm$5.8 & 76.4$\pm$6.5 & 81.6$\pm$5.0 \\
    VideoMAE~\cite{tong2022videomae} & RGB & 82.3$\pm$3.0 & 79.5$\pm$5.3 & 83.9$\pm$4.5 & 79.6$\pm$3.9 & 83.4$\pm$3.7 \\
    \midrule
    \rowcolor{gray!20} \multicolumn{7}{l}{\textbf{Medical Stroke Diagnosis Methods}} \\
    DeepStroke~\cite{CAI2022102522} & Multi & 85.3$\pm$5.3 & 77.4$\pm$5.6 & 83.5$\pm$4.5 & 83.2$\pm$4.2 & 86.0$\pm$3.6 \\
    \rowcolor{gray!8} M3Stroke~\cite{cai2024m3stroke} & Multi & \underline{87.1$\pm$4.1} & 80.5$\pm$3.7 & 85.4$\pm$4.0 & \underline{85.2$\pm$3.4} & \underline{88.0$\pm$3.0} \\
    \midrule
    \rowcolor{gray!20} \multicolumn{7}{l}{\textbf{Audio-only Methods (speech analysis)}} \\
    wav2vec 2.0~\cite{baevski2020wav2vec2} & Audio & 64.3$\pm$3.5 & 72.4$\pm$4.5 & 74.6$\pm$6.5 & 61.0$\pm$5.0 & 76.1$\pm$4.7 \\
    \rowcolor{gray!8} Whisper-base~\cite{radford2023robust} & Audio & 67.0$\pm$4.0 & 70.5$\pm$4.9 & 72.4$\pm$5.7 & 64.1$\pm$4.6 & 73.3$\pm$5.1 \\
    WavLM~\cite{chen2022wavlm} & Audio & 69.6$\pm$3.6 & 73.9$\pm$4.1 & 76.2$\pm$5.0 & 67.4$\pm$3.9 & 77.5$\pm$3.5 \\
    \rowcolor{gray!8} HuBERT~\cite{9585401} & Audio & 68.4$\pm$3.9 & 73.0$\pm$4.4 & 75.3$\pm$5.3 & 65.9$\pm$4.3 & 76.0$\pm$4.0 \\
    \midrule
    \rowcolor{gray!20} \multicolumn{7}{l}{\textbf{Architectural Ablations (using our features)}} \\
    Late Fusion & Multi & 83.5$\pm$4.0 & 79.4$\pm$4.9 & 84.4$\pm$3.6 & 81.0$\pm$3.7 & 84.3$\pm$3.3 \\
    \rowcolor{gray!8} Concatenation & Multi & 85.8$\pm$2.9 & 81.7$\pm$4.1 & 86.1$\pm$2.9 & 83.5$\pm$3.2 & 86.4$\pm$2.7 \\
    Cross-Attention & Multi & 89.8$\pm$2.0 & \underline{84.3$\pm$4.5} & \underline{88.4$\pm$3.2} & 87.6$\pm$2.6 & 90.2$\pm$1.9 \\
    \midrule
    \textbf{FAST-CAD (Ours)} & Multi & \textbf{92.5$\pm$1.3} & \textbf{88.6$\pm$2.8} & \textbf{91.9$\pm$2.1} & \textbf{90.4$\pm$1.8} & \textbf{93.1$\pm$1.6} \\
    \bottomrule
  \end{tabular}
\end{table*}

\subsection{Complete SOTA Comparison}

Table~\ref{tab:full_sota} presents the comprehensive comparison with all evaluated methods, including detailed sensitivity and specificity metrics crucial for clinical deployment.

\subsubsection{Detailed Analysis}

\textbf{Audio-only Performance:} Audio-only methods achieve limited performance (64-70\% AUC) with notably low sensitivity (61.0-67.4\%), as speech impairments occur in only ~40\% of stroke cases. However, they maintain reasonable specificity (73.3-77.5\%), indicating they rarely misclassify healthy individuals as stroke patients.

\textbf{Key Performance Insights:}
\begin{itemize}[leftmargin=*, itemsep=1pt, topsep=2pt]
\item \textbf{vs. Best Medical Method (M3Stroke):} +5.4\% AUC, +8.1\% Acc, +5.2\% Sens, +5.1\% Spec
\item \textbf{vs. Best Video Method (VideoMAE):} +10.2\% AUC, +9.1\% Acc, +10.8\% Sens, +9.7\% Spec
\item \textbf{vs. Best Ablation (Cross-Attention):} +2.7\% AUC, +4.3\% Acc, +2.8\% Sens, +2.9\% Spec
\end{itemize}

\textbf{Clinical Relevance:} FAST-CAD achieves 90.4\% sensitivity and 93.1\% specificity, crucial for clinical acceptance. High sensitivity ensures stroke patients are not missed (critical for patient safety), while high specificity reduces false alarms (important for healthcare resource allocation).

\subsection{Full Modality Ablation Study}
\label{sec:full_modality}

Table~\ref{tab:full_modality} presents a comprehensive ablation study of all possible modality combinations, demonstrating the contribution of each modality and their synergistic effects.

\begin{table*}[ht]
  \small
  \caption{Complete modality ablation study showing all combinations (F=Face, T=Tongue, B=Body motion, A=Audio). Results show mean $\pm$ standard deviation over 5 runs. $\Delta$ shows improvement over face-only baseline.}
  \label{tab:full_modality}
  \centering
  \begin{tabular}{@{}lccccccc@{}}
    \toprule
    \textbf{Modalities} & \textbf{AUC (\%)} & \textbf{$\Delta$AUC} & \textbf{Acc (\%)} & \textbf{F1 (\%)} & \textbf{Sens (\%)} & \textbf{Spec (\%)} & \textbf{Params} \\
    \midrule
    \rowcolor{gray!20} \multicolumn{8}{l}{\textbf{Single Modality}} \\
    F only & 83.4$\pm$4.9 & -- & 78.6$\pm$6.0 & 82.4$\pm$5.2 & 80.3$\pm$5.6 & 84.4$\pm$4.7 & 28M \\
    \rowcolor{gray!8} T only & 74.6$\pm$6.6 & -8.8 & 70.4$\pm$7.1 & 73.3$\pm$6.7 & 71.5$\pm$5.9 & 75.4$\pm$5.5 & 21M \\
    B only & 71.4$\pm$7.0 & -12.0 & 68.1$\pm$7.7 & 70.5$\pm$7.2 & 69.3$\pm$6.6 & 72.5$\pm$6.0 & 17M \\
    \rowcolor{gray!8} A only & 68.4$\pm$3.9 & -15.0 & 73.0$\pm$4.4 & 75.3$\pm$5.3 & 65.9$\pm$4.3 & 76.0$\pm$4.0 & 14M \\
    \midrule
    \rowcolor{gray!20} \multicolumn{8}{l}{\textbf{Two Modalities}} \\
    F+T & 87.0$\pm$4.5 & +3.6 & 82.2$\pm$5.2 & 85.5$\pm$4.6 & 83.9$\pm$4.9 & 88.0$\pm$4.1 & 38M \\
    \rowcolor{gray!8} F+B & 85.4$\pm$5.1 & +2.0 & 80.7$\pm$5.6 & 84.0$\pm$5.0 & 82.4$\pm$5.2 & 86.5$\pm$4.5 & 35M \\
    F+A & 85.1$\pm$4.7 & +1.7 & 80.4$\pm$5.3 & 83.7$\pm$4.7 & 82.1$\pm$5.0 & 86.2$\pm$4.3 & 33M \\
    \rowcolor{gray!8} T+B & 78.0$\pm$6.0 & -5.4 & 73.6$\pm$6.5 & 76.4$\pm$6.1 & 74.6$\pm$5.7 & 79.0$\pm$5.2 & 29M \\
    T+A & 76.5$\pm$5.6 & -6.9 & 72.5$\pm$6.1 & 75.2$\pm$5.7 & 73.3$\pm$5.3 & 77.5$\pm$4.9 & 27M \\
    \rowcolor{gray!8} B+A & 74.1$\pm$6.3 & -9.3 & 70.0$\pm$6.8 & 72.6$\pm$6.4 & 71.0$\pm$6.0 & 75.1$\pm$5.6 & 24M \\
    \midrule
    \rowcolor{gray!20} \multicolumn{8}{l}{\textbf{Three Modalities}} \\
    F+T+B & 89.5$\pm$2.0 & +6.1 & 84.8$\pm$3.6 & 88.1$\pm$2.9 & 86.6$\pm$3.3 & 90.5$\pm$2.7 & 45M \\
    \rowcolor{gray!8} F+T+A & 88.3$\pm$3.2 & +4.9 & 83.6$\pm$4.0 & 86.9$\pm$3.5 & 85.4$\pm$3.7 & 89.3$\pm$3.1 & 43M \\
    F+B+A & 87.0$\pm$3.6 & +3.6 & 82.4$\pm$4.3 & 85.7$\pm$3.8 & 84.2$\pm$4.0 & 88.1$\pm$3.4 & 40M \\
    \rowcolor{gray!8} T+B+A & 79.6$\pm$5.4 & -3.8 & 75.5$\pm$5.9 & 78.2$\pm$5.5 & 76.5$\pm$5.1 & 80.7$\pm$4.7 & 35M \\
    \midrule
    \rowcolor{gray!20} \multicolumn{8}{l}{\textbf{All Modalities}} \\
    \textbf{F+T+B+A} & \textbf{92.5$\pm$1.3} & \textbf{+9.1} & \textbf{88.6$\pm$2.8} & \textbf{91.9$\pm$2.1} & \textbf{90.4$\pm$1.8} & \textbf{93.1$\pm$1.6} & 59M \\
    \bottomrule
  \end{tabular}
\end{table*}

\textbf{Key Insights from Modality Ablation:}
\begin{itemize}[leftmargin=*, itemsep=1pt, topsep=2pt]
\item \textbf{Face modality is crucial:} All top-performing combinations include face (F), achieving $>$83\% AUC. Without face, performance drops significantly (-3.8 to -15.0\% AUC).
\item \textbf{Synergistic effects:} Combining modalities yields super-additive gains. F+T+B (89.5\%) outperforms the sum of individual improvements from T (+3.6\%) and B (+2.0\%).
\item \textbf{Audio complements visual:} Adding audio to F+T+B provides the final +3.0\% AUC boost, capturing speech impairments missed by visual modalities.
\item \textbf{Diminishing returns:} Each additional modality provides smaller incremental gains, but all contribute to reducing variance (from 4.9\% to 1.3\%).
\end{itemize}

\subsection{Detailed Architectural Ablation}
\label{sec:arch_ablation}

\begin{table*}[ht]
\centering
\caption{Complete cross-domain generalization performance. All methods evaluated on 860-participant external cohort collected under different conditions (consumer cameras, home settings, telemedicine protocol).}
\label{tab:detailed_generalization}
\small
\begin{tabular}{@{}l|cccc|cccc@{}}
\toprule
\multirow{2}{*}{\textbf{Method}} & \multicolumn{4}{c|}{\textbf{Original Test Set}} & \multicolumn{4}{c}{\textbf{External Cohort}} \\
\cmidrule{2-9}
& AUC (\%) & Acc (\%) & Sens (\%) & Spec (\%) & AUC (\%) & Acc (\%) & Sens (\%) & Spec (\%) \\
\midrule
\rowcolor{gray!8} MViT & 79.2$\pm$8.7 & 82.3$\pm$6.8 & 76.4$\pm$6.5 & 81.6$\pm$5.0 & 66.5$\pm$10.9 & 68.6$\pm$9.5 & 63.0$\pm$10.1 & 72.4$\pm$8.3 \\
M3Stroke & 87.1$\pm$4.1 & 80.5$\pm$3.7 & 85.2$\pm$3.4 & 88.0$\pm$3.0 & 72.8$\pm$7.5 & 69.7$\pm$6.5 & 68.1$\pm$7.9 & 75.5$\pm$6.8 \\
\rowcolor{gray!8} DeepStroke & 85.3$\pm$5.3 & 77.4$\pm$5.6 & 83.4$\pm$4.5 & 86.0$\pm$3.6 & 68.0$\pm$9.0 & 66.0$\pm$7.9 & 64.4$\pm$9.4 & 71.0$\pm$8.3 \\
\midrule
\textbf{FAST-CAD (Full)} & \textbf{92.5$\pm$1.3} & \textbf{88.6$\pm$2.8} & \textbf{90.4$\pm$1.8} & \textbf{93.1$\pm$1.6} & \textbf{85.1$\pm$2.6} & \textbf{80.3$\pm$3.5} & \textbf{82.8$\pm$3.0} & \textbf{87.1$\pm$2.7} \\
\bottomrule
\end{tabular}
\vspace{4pt}
\caption{Complete architectural component ablation comparing different implementation choices and their impact on performance.}
\label{tab:detailed_arch}
\begin{tabular}{@{}lccccc@{}}
    \toprule
    \textbf{Configuration} & \textbf{AUC (\%)} & \textbf{Acc (\%)} & \textbf{F1 (\%)} & \textbf{Sens (\%)} & \textbf{Spec (\%)} \\
    \midrule
    Full Model (Ours) & \textbf{92.5$\pm$1.3} & \textbf{88.6$\pm$2.8} & \textbf{91.9$\pm$2.1} & \textbf{90.4$\pm$1.8} & \textbf{93.1$\pm$1.6} \\
    \midrule
    \rowcolor{gray!20} \multicolumn{6}{l}{\textbf{Encoder Variants}} \\
    \rowcolor{gray!8} SeCo → VideoMAE & 85.5$\pm$2.8 & 84.8$\pm$3.8 & 88.3$\pm$3.0 & 83.3$\pm$3.6 & 86.4$\pm$3.3 \\
    HuBERT → VGGish & 87.8$\pm$3.3 & 84.1$\pm$4.0 & 84.5$\pm$3.5 & 85.4$\pm$2.9 & 89.2$\pm$2.7 \\
    \rowcolor{gray!8} w/o Pre-trained & 86.6$\pm$3.0 & 81.2$\pm$3.9 & 85.3$\pm$3.5 & 84.3$\pm$3.2 & 87.5$\pm$3.0 \\
    \midrule
    \rowcolor{gray!20} \multicolumn{6}{l}{\textbf{Fusion Mechanisms}} \\
    Alternating Dual-Stream & \textbf{92.5$\pm$1.3} & \textbf{88.6$\pm$2.8} & \textbf{91.9$\pm$2.1} & \textbf{90.4$\pm$1.8} & \textbf{93.1$\pm$1.6} \\
    \rowcolor{gray!8} Single Cross-Attention & 89.8$\pm$2.0 & 84.3$\pm$4.5 & 88.4$\pm$3.2 & 87.6$\pm$2.6 & 90.2$\pm$1.9 \\
    Simple Concatenation & 85.8$\pm$2.9 & 81.7$\pm$4.1 & 86.1$\pm$2.9 & 83.5$\pm$3.2 & 86.4$\pm$2.7 \\
    \rowcolor{gray!8} Late Fusion & 83.5$\pm$4.0 & 79.4$\pm$4.9 & 84.4$\pm$3.6 & 81.0$\pm$3.7 & 84.3$\pm$3.3 \\
    \midrule
    \rowcolor{gray!20} \multicolumn{6}{l}{\textbf{Auxiliary Components}} \\
    w/o Keypoint Branch & 90.4$\pm$2.5 & 84.7$\pm$3.2 & 88.5$\pm$2.6 & 88.0$\pm$2.3 & 91.2$\pm$2.0 \\
    \rowcolor{gray!8} w/o Adversarial Disc. & 89.3$\pm$3.0 & 85.3$\pm$3.7 & 87.9$\pm$3.3 & 86.5$\pm$2.9 & 90.0$\pm$2.5 \\
    \bottomrule
  \end{tabular}
\end{table*}

Table~\ref{tab:detailed_arch} provides comprehensive architectural component analysis comparing different implementation choices and their impact on performance.

\textbf{Detailed Analysis:}
\begin{itemize}[leftmargin=*, itemsep=1pt, topsep=2pt]
\item \textbf{Encoder Impact:} SeCo provides superior video understanding compared to VideoMAE (+7.0\% AUC), while HuBERT's speech-specific pretraining outperforms general audio models like VGGish (+4.7\% AUC).
\item \textbf{Fusion Architecture:} Alternating dual-stream enables bidirectional cross-modal information exchange, significantly outperforming single cross-attention (+2.7\% AUC) and simple concatenation (+6.7\% AUC).
\item \textbf{Component Contribution:} Keypoint branch provides structured motion guidance (+2.1\% AUC), while adversarial discriminator ensures demographic invariance (+3.2\% AUC).
\end{itemize}

\subsection{Complete Cross-Domain Analysis}
\label{sec:cross_domain}

Table~\ref{tab:detailed_generalization} presents comprehensive cross-domain evaluation results on the external validation cohort.

\textbf{Key Cross-Domain Insights:}
\begin{itemize}[leftmargin=*, itemsep=1pt, topsep=2pt]
\item \textbf{Superior Robustness:} FAST-CAD maintains smallest performance drop (7.4\% AUC vs. 12.7-14.3\% for baselines).
\item \textbf{Clinical Safety:} Maintains clinically acceptable sensitivity (82.8\%) and specificity (87.1\%) even under domain shift.
\item \textbf{Fairness Preservation:} Worst-group AUC remains at 81.9\% (vs. 65.4\% for M3Stroke), demonstrating robust fairness guarantees.
\end{itemize}

\section{Detailed Theoretical Proofs}
\label{sec:theory_proofs}

\raggedbottom
\setlength{\abovedisplayskip}{3pt}
\setlength{\belowdisplayskip}{3pt}
\setlength{\abovedisplayshortskip}{1pt}
\setlength{\belowdisplayshortskip}{1pt}
\setlength{\parskip}{0pt}

\noindent\textbf{Domain-Adversarial Training for Fairness}\quad
We extend Ben-David et al.'s domain adaptation theory~\cite{ben2010theory} to the fairness setting by treating each demographic group as a distinct domain.
\textbf{Theorem (Fairness Bound):} For any hypothesis $h \in \mathcal{H}$ and groups $g_i, g_j \in \mathcal{G}$, the performance gap is bounded by:
\begin{equation}
|R_{g_i}(h) - R_{g_j}(h)| \leq d_{\mathcal{H}\Delta\mathcal{H}}(P_{g_i}, P_{g_j}) + \lambda_{ij}^*
\end{equation}
where $R_g(h) = \mathbb{E}_{(\mathbf{x},y) \sim P_g}[\ell(h(\mathbf{x}), y)]$ is the group-specific risk, $d_{\mathcal{H}\Delta\mathcal{H}}$ measures distributional discrepancy, and $\lambda_{ij}^* = \min_{h \in \mathcal{H}}[R_{g_i}(h) + R_{g_j}(h)]$ represents the irreducible error.
To minimize $d_{\mathcal{H}\Delta\mathcal{H}}$ across all group pairs, we employ domain discriminators $D_{\xi_k}: \mathbb{R}^d \to \mathcal{A}_k$ for each demographic attribute $k$. The gradient reversal layer (GRL) implements adversarial training:
\begin{equation}
\operatorname{GRL}_{\lambda}(\mathbf z) = \mathbf z \;\text{(forward)}, \quad \frac{\partial \operatorname{GRL}_{\lambda}}{\partial \mathbf z} = -\lambda\,\mathbf I \;\text{(backward)}
\end{equation}
The adversarial loss for attribute $k$ is:
\begin{equation}
\mathcal{L}_{\text{adv}}^k(\theta,\xi_k) = \mathbb{E}_{(\mathbf x,\mathbf a)\sim P}[\operatorname{CE}(D_{\xi_k}(\operatorname{GRL}_{\lambda}(h_{\psi}(g_{\phi}(\mathbf x)))), a_k)]
\end{equation}
This formulation directly connects to fairness: minimizing $\mathcal{L}_{\text{adv}}^k$ reduces the mutual information $I(h_\psi(g_\phi(\mathbf{X})); A_k)$, thereby decreasing demographic discriminability in the learned representations.

\noindent\textbf{Group-DRO Convergence Analysis}\quad
Group-DRO addresses worst-case performance across demographic subgroups by solving:
\begin{equation}
\min_\theta \max_{g \in \mathcal{G}} R_g(\theta), \quad \text{where} \quad R_g(\theta) = \mathbb{E}_{(\mathbf{x},y) \sim P_g} [\ell(f_\theta(\mathbf{x}), y)]
\end{equation}
Following Sagawa et al.~\cite{sagawa2020distributionally}, we maintain importance weights $q = (q_1, \ldots, q_G) \in \Delta^{G-1}$ over groups, updated via exponentiated gradient:
\begin{equation}
q_g^{(t+1)} = \frac{q_g^{(t)} \exp(\eta \hat{R}_g^{(t)})}{\sum_{j=1}^G q_j^{(t)} \exp(\eta \hat{R}_j^{(t)})}, \quad \eta = \sqrt{\frac{\log G}{T}}
\end{equation}
\textbf{Convergence Guarantee:} Under standard assumptions (L-Lipschitz loss, bounded gradients), the algorithm achieves $O(\sqrt{\log G / T})$ convergence to the minimax solution with high probability.

\noindent\textbf{Unified Framework: Synergy between DAT and Group-DRO}\quad
We establish the theoretical connection between domain-adversarial training and Group-DRO through representation discriminability. Let $Z = h_\psi(g_\phi(\mathbf{X}))$ be the learned representation.
\textbf{Theorem (Fairness-Performance Trade-off):} Under mild regularity conditions, for any classifier $c_\omega$ and demographic attribute $k$:
\begin{equation}
\max_{g \in \mathcal{G}} R_g(c_\omega \circ h_\psi \circ g_\phi) \leq R_{\text{avg}}(c_\omega \circ h_\psi \circ g_\phi) + \beta \sqrt{I(Z; A_k)} + \gamma
\end{equation}
where $R_{\text{avg}} = \mathbb{E}_g[R_g]$ is the average risk, $\beta$ depends on the Lipschitz constant of the loss, and $\gamma$ captures irreducible group differences.

This theorem reveals the synergistic effect: (i) DAT minimizes $I(Z; A_k)$ through adversarial training, reducing the bound's second term; (ii) Group-DRO directly optimizes the left-hand side; (iii) Their combination provides both theoretical guarantees and practical performance.

\noindent\textbf{Notation and Distance Measures}\quad
For consistency throughout the proofs, we define the following distance measures and their scaling:
  \begin{itemize}[
      leftmargin=*,
      labelsep=0.4em,
      itemsep=3pt,   
      topsep=2pt     
    ]
    \item \textbf{Pinsker's inequality:}\;
      \[d_{\mathrm{TV}}(P,Q)\le\sqrt{\tfrac12\,D_{KL}(P\!\parallel\!Q)}\]
      \textit{(tight for binary distributions)}

    \item \textbf{Jensen--Shannon bound:}\;
      \[JS(P,Q)\le\tfrac12\,d_{\mathrm{TV}}(P,Q)\]
      \textit{(constant is tight)}

    \item \textbf{Bobkov--G\"otze (sub-Gaussian):}\;
      \[W_{1}(P,Q)\le\sqrt{2\pi}\,\sigma\sqrt{d}\,
      \sqrt{D_{KL}(P\!\parallel\!Q)}\]
      \textit{($\sigma^2$ = sub-Gaussian parameter, constant optimal)}

    \item \textbf{Bounded distributions \(\bigl([-B,B]^d\bigr)\):}\;
      \[W_{1}(P,Q)\le 2B\sqrt{d}\,
      \sqrt{D_{KL}(P\!\parallel\!Q)}\]
      \textit{(constant tight on the cube)}

    \item \textbf{Hoeffding's inequality:}\;
      \[\Pr\!\bigl[\lvert S_{n}-\mathbb{E}[S_{n}]\rvert>t\bigr]
      \le 2\exp\!\bigl(-\tfrac{2n t^{2}}{(b-a)^{2}}\bigr)\]
      \textit{(exact for $[a,b]$-bounded r.v.)}
  \end{itemize}

\begin{table}[ht]
  \small
  \caption{Notation used throughout the paper.}
  \label{tab:notation}
  \centering
  \resizebox{\columnwidth}{!}{%
    \begin{tabular}{@{}l l@{}}
      \toprule
      \textbf{Notation} & \textbf{Definition} \\
      \midrule
      $d_{\mathcal{H}\Delta\mathcal{H}}(S,T)$ & $2\!\sup_{h,h' \in \mathcal{H}}\!\bigl|\,\Pr_{S}[h \neq h'] - \Pr_{T}[h \neq h']\,\bigr|$ (with factor 2) \\
      $d_{TV}(P,Q)$                           & $\tfrac12\int |p(x) - q(x)|\,dx$ (total variation) \\
      $W_{1}(P,Q)$                            & $\inf_{\pi} \mathbb{E}_{(X,Y)\sim\pi}\!\left[ \lVert X-Y\rVert \right]$ (Wasserstein-1) \\
      $JS(P,Q)$                               & $\tfrac12 KL\!\bigl(P \,\|\,\tfrac{P+Q}{2}\bigr)+\tfrac12 KL\!\bigl(Q \,\|\,\tfrac{P+Q}{2}\bigr)$ (Jensen–Shannon) \\
      $D_{KL}(P\|Q)$                          & $\int p(x)\,\log\!\frac{p(x)}{q(x)}\,dx$ (Kullback–Leibler) \\
      $\sigma^{2}$                            & Sub-Gaussian parameter (for bounded $[0,1]$ losses: $\sigma^{2}\!\le\!1/4$, tight) \\
      \bottomrule
    \end{tabular}%
  }
\end{table}

\begin{proof}

\textbf{Step 1: Triangle Inequality with Joint Optimal Hypothesis.}
Let $h_{ST}^* = \arg\min_{h \in \mathcal{H}} [\varepsilon_S(h) + \varepsilon_T(h)]$ be the hypothesis minimizing joint error. For any $h \in \mathcal{H}$:
\begin{align}
\varepsilon_T(h)
  &= \mathbb{E}_{(x,y)\sim T}\!\bigl[\mathbf{1}_{\{h(x)\neq y\}}\bigr] \\[2pt]
  &\le \mathbb{E}_{x\sim T}\!\bigl[\mathbf{1}_{\{h(x)\neq h_{ST}^*(x)\}}\bigr]
       \notag\\
  &\quad + \mathbb{E}_{(x,y)\sim T}\!\bigl[\mathbf{1}_{\{h_{ST}^*(x)\neq y\}}\bigr] \\[2pt]
  &= d_T\!\bigl(h,h_{ST}^*\bigr) + \varepsilon_T\!\bigl(h_{ST}^*\bigr).
\end{align}

This avoids the realizability assumption since $h_{ST}^* \in \mathcal{H}$ by construction.

\textbf{Step 2: Domain Discrepancy Connection.}
The key insight is to relate $d_T(h, h_{ST}^*)$ to $d_S(h, h_{ST}^*)$ via the $\mathcal{H}\Delta\mathcal{H}$-distance:

\begin{align}
d_T(h,h_{ST}^*)
  &= d_S(h,h_{ST}^*)
     + \bigl[d_T(h,h_{ST}^*)-d_S(h,h_{ST}^*)\bigr]\notag\\
  &\le d_S(h,h_{ST}^*)
     + \bigl|d_T(h,h_{ST}^*)-d_S(h,h_{ST}^*)\bigr|\notag\\
  &\le d_S(h,h_{ST}^*)
  \notag\\
  &\quad
    + \sup_{h_1,h_2\in\mathcal H}\bigl|d_T(h_1,h_2)-d_S(h_1,h_2)\bigr|.
\end{align}

Note that $\sup_{h_1,h_2 \in \mathcal{H}} |d_T(h_1, h_2) - d_S(h_1, h_2)| = \frac{1}{2}d_{\mathcal{H}\Delta\mathcal{H}}(S,T)$ because:
- $d_T(h_1, h_2) = \Pr_{x \sim T}[h_1(x) \neq h_2(x)]$
- $d_{\mathcal{H}\Delta\mathcal{H}}(S,T)$ contains the factor 2 in its definition
- Therefore: $d_T(h, h_{ST}^*) \leq d_S(h, h_{ST}^*) + \frac{1}{2}d_{\mathcal{H}\Delta\mathcal{H}}(S,T)$

\textbf{Step 3: Source Error Decomposition.}
For the source domain distance, we use the fundamental relationship:
$$d_S(h, h_{ST}^*) \leq \varepsilon_S(h) + \varepsilon_S(h_{ST}^*)$$

This follows from the triangle inequality: $\mathbbm{1}[h(x) \neq h_{ST}^*(x)] \leq \mathbbm{1}[h(x) \neq y] + \mathbbm{1}[y \neq h_{ST}^*(x)]$ and taking expectation.

\textbf{Step 4: Final Bound Assembly.}
Combining steps 1-3:
\begin{align}
\varepsilon_T(h) &\leq d_T(h, h_{ST}^*) + \varepsilon_T(h_{ST}^*) \\
&\leq d_S(h, h_{ST}^*) + \frac{1}{2}d_{\mathcal{H}\Delta\mathcal{H}}(S,T) + \varepsilon_T(h_{ST}^*) \\
&\leq \varepsilon_S(h) + \varepsilon_S(h_{ST}^*) + \frac{1}{2}d_{\mathcal{H}\Delta\mathcal{H}}(S,T) + \varepsilon_T(h_{ST}^*) \\
&= \varepsilon_S(h) + [\varepsilon_S(h_{ST}^*) + \varepsilon_T(h_{ST}^*)] + \frac{1}{2}d_{\mathcal{H}\Delta\mathcal{H}}(S,T) \\
&= \varepsilon_S(h) + \lambda^* + \frac{1}{2}d_{\mathcal{H}\Delta\mathcal{H}}(S,T)
\end{align}

where the last equality uses $h_{ST}^* = \arg\min_{h \in \mathcal{H}} [\varepsilon_S(h) + \varepsilon_T(h)]$, so $\varepsilon_S(h_{ST}^*) + \varepsilon_T(h_{ST}^*) = \lambda^*$ by definition.
Note: In the standard Ben-David et al. (2010) formulation, the $\mathcal{H}\Delta\mathcal{H}$-distance is defined as:
\begin{align}
d_{\mathcal H\Delta\mathcal H}(S,T)
  &= 2 \sup_{h,h' \in \mathcal H}
     \Bigl|
       \Pr_{x \sim S}\!\bigl[h(x) \neq h'(x)\bigr] \notag\\
  &\quad - \Pr_{x \sim T}\!\bigl[h(x) \neq h'(x)\bigr]
     \Bigr|.
     \label{eq:h-delta-h}
\end{align}
The factor of 2 in this definition leads to the factor of $\frac{1}{2}$ when the bound is expressed in standard form.
\end{proof}

\subsection{Proof of DANN Gradient Reversal Convergence}

\begin{proof}[DANN Convergence Analysis]

Consider the domain-adversarial training objective where the feature encoder $G_f$ and domain discriminator $D$ engage in a minimax game:
\begin{align}
\min_{G_f}\;\max_{D}\;
\Bigl[
  \mathcal{L}_{\mathrm{task}}(G_f)
  &\;+\;\lambda\,\mathcal{L}_{\mathrm{disc}}(G_f,D)
\Bigr]
\label{eq:dat-minimax} \\
\mathcal{L}_{\mathrm{disc}}(G_f,D)
&= -\mathbb{E}_{x\sim S}\bigl[\log D(G_f(x))\bigr]\notag\\
 &\quad  -\mathbb{E}_{x\sim T}\bigl[\log\bigl(1 - D(G_f(x))\bigr)\bigr]
\label{eq:disc-loss}
\end{align}

\textbf{Step 1: Saddle Point Characterization.}
At equilibrium, the discriminator $D^*$ achieves:
\begin{align}
D^*
  &= \arg\max_{D}\Bigl\{
       \mathbb{E}_{x\sim S}\bigl[\log D(G_f(x))\bigr] \notag\\
  &\quad + \mathbb{E}_{x\sim T}\bigl[\log\bigl(1 - D(G_f(x))\bigr)\bigr]
     \Bigr\}.
\label{eq:opt-disc-align}
\end{align}

The optimal discriminator satisfies:
$$D^*(z) = \frac{p_S(z)}{p_S(z) + p_T(z)}$$
where $p_S(z)$ and $p_T(z)$ are densities of encoded features from source and target domains.

\textbf{Step 2: Domain Discrepancy Connection.}
The classification accuracy of the optimal discriminator is:
\begin{align}
\text{acc}(D^*)
  &= \tfrac12 \!\int\! p_S(z)\,D^*(z)\,dz
  \notag\\
  &\quad + \tfrac12 \!\int\! p_T(z)\!\bigl(1\!-\!D^*(z)\bigr)\,dz \notag\\
  &= \tfrac14 \!\int
     \frac{p_S(z)^2 + p_T(z)^2}{p_S(z)+p_T(z)}\,dz
  \notag\\
  &\quad + \tfrac14 \!\int \bigl(p_S(z)+p_T(z)\bigr)\,dz \notag\\
  &= \tfrac12
     + \tfrac14 \!\int
     \frac{\bigl(p_S(z)\!-\!p_T(z)\bigr)^2}{p_S(z)+p_T(z)}\,dz.
\end{align}

The optimal discriminator loss (using natural logarithm) is:
$$\mathcal{L}_{\text{disc}}^* = -\ln(4) + 2 \cdot JS(p_S, p_T)$$
where the Jensen-Shannon divergence is:
$$JS(p_S, p_T) = \frac{1}{2}KL(p_S \| \frac{p_S + p_T}{2}) + \frac{1}{2}KL(p_T \| \frac{p_S + p_T}{2})$$

\textbf{Connection to $\mathcal{H}\Delta\mathcal{H}$-distance:}
Under the assumption that the discriminator class is sufficiently expressive to approximate $\mathcal{H}\Delta\mathcal{H}$ and assuming linear classifiers with rich features, we have the approximate relationship:
$$d_{\mathcal{H}\Delta\mathcal{H}}(S,T) \approx C \cdot 2(\text{acc}(D^*) - \frac{1}{2})$$
where $C$ is a problem-dependent constant that depends on the expressiveness of the discriminator class relative to $\mathcal{H}$.

\textbf{Step 3: Two-Timescale Convergence Analysis.}
\textbf{Assumptions (Robbins-Monro conditions):}
\begin{itemize}
\item[\textbf{A1}] $G_f$ and $D$ have $L$-Lipschitz gradients: $\|\nabla_f \mathcal{L}(f_1, D) - \nabla_f \mathcal{L}(f_2, D)\| \leq L\|f_1 - f_2\|$
\item[\textbf{A2}] Loss functions are bounded: $|\mathcal{L}_{\text{task}}|, |\mathcal{L}_{\text{disc}}| \leq B$
\item[\textbf{A3}] Gradient norms are bounded: $\|\nabla_f \mathcal{L}\| \leq G_f$, $\|\nabla_D \mathcal{L}\| \leq G_D$
\item[\textbf{A4}] Learning rates satisfy: $\sum_t \eta_f^{(t)} = \infty$, $\sum_t (\eta_f^{(t)})^2 < \infty$, and $\eta_d^{(t)} = o(\eta_f^{(t)})$
\item[\textbf{A5}] Stochastic gradients have bounded second moments
\end{itemize}

Using stochastic gradient descent-ascent:
\begin{align}
f^{(t+1)} &= f^{(t)} - \eta_f^{(t)} [\nabla_f \mathcal{L}(f^{(t)}, D^{(t)}) + \xi_f^{(t)}] \\
D^{(t+1)} &= D^{(t)} + \eta_d^{(t)} [\nabla_D \mathcal{L}(f^{(t)}, D^{(t)}) + \xi_D^{(t)}]
\end{align}
where $\xi_f^{(t)}, \xi_D^{(t)}$ are zero-mean noise terms.

By Borkar (2008, Theorem 2.1, p. 45) on two-timescale stochastic approximation, under assumptions A1-A5 (Robbins-Monro conditions), the iterates converge almost surely to the saddle point $(f^*, D^*)$. The mean-square convergence rate is:
$$\mathbb{E}[\|f^{(t)} - f^*\|^2 + \|D^{(t)} - D^*\|^2] \leq \frac{C}{\min(t^{1/3}, (\eta_f^{(t)})^{-1})}$$

For constant learning rates $\eta_f = O(t^{-2/3})$ and $\eta_d = O(t^{-1})$, this yields $O(t^{-1/3})$ convergence to the equilibrium where $d_{\mathcal{H}\Delta\mathcal{H}} \to 0$.
\end{proof}

\subsection{Proof of Group-DRO Convergence Rate}

\begin{proof}[Group-DRO Convergence Analysis]
We analyze the convergence of the exponential weights algorithm for Group-DRO with learning rate $\eta$.

\textbf{Step 1: Regret Decomposition.}
Define the regret against the best fixed group weighting:
$$\text{Regret}_T = \sum_{t=1}^T \sum_{g=1}^G q_g^{(t)} R_g^{(t)} - \min_{q \in \Delta_G} \sum_{t=1}^T \sum_{g=1}^G q_g R_g^{(t)}$$

where $\Delta_G$ is the probability simplex over $G$ groups.

\textbf{Step 2: Exponential Weights Algorithm.}
The weight update follows the exponential weights (multiplicative weights) rule:
$$q_g^{(t+1)} = \frac{q_g^{(t)} \exp(\eta R_g^{(t)})}{\sum_{j=1}^G q_j^{(t)} \exp(\eta R_j^{(t)})}$$

This is equivalent to mirror descent with the negative entropy regularizer $\Psi(q) = \sum_g q_g \log q_g$.

\textbf{Step 3: Regret Analysis via Relative Entropy.}
Define the relative entropy (KL divergence) between any distribution $q$ and the current weights:
$$D_{KL}(q \| q^{(t)}) = \sum_{g=1}^G q_g \log\frac{q_g}{q_g^{(t)}}$$

By the standard analysis of exponential weights (Arora et al., 2012), for any comparison distribution $q^*$:
\begin{align}
\sum_{t=1}^T \langle q^{(t)} - q^*, R^{(t)} \rangle &\leq \frac{D_{KL}(q^* \| q^{(1)})}{\eta} \notag\\
&\quad + \eta \sum_{t=1}^T \sum_{g=1}^G q_g^{(t)} (R_g^{(t)})^2
\end{align}

With uniform initialization $q^{(1)} = \frac{1}{G}\mathbf{1}$, we have $D_{KL}(q^* \| q^{(1)}) \leq \log G$.

\textbf{Step 4: Regret Upper Bound.}
Under the assumption that losses are bounded $|R_g^{(t)}| \leq 1$ for all $g,t$, we have:
$$\sum_{g=1}^G q_g^{(t)} (R_g^{(t)})^2 \leq \max_g |R_g^{(t)}|^2 \leq 1$$

Therefore, the regret bound becomes:
\begin{align}
\text{Regret}_T &= \max_{q^* \in \Delta_G} \sum_{t=1}^T \langle q^{(t)} - q^*, R^{(t)} \rangle \\
&\leq \frac{\log G}{\eta} + \eta T
\end{align}

\textbf{Step 5: Optimal Learning Rate and Convergence.}
To minimize the bound $\frac{\log G}{\eta} + \eta T$, we take the derivative and set it to zero:
$$-\frac{\log G}{\eta^2} + T = 0 \Rightarrow \eta^* = \sqrt{\frac{\log G}{T}}$$

Substituting back:
$$\text{Regret}_T \leq 2\sqrt{T \log G}$$

For symmetric losses in $[0,1]$, the optimal constant is $2\sqrt{2}$ (Shalev-Shwartz, 2012, Section 2.3). Therefore:
$$\text{Regret}_T \leq 2\sqrt{2T \log G}$$

\textbf{Empirical Validation:} In our experiments with $G=12$ groups, we observe:
\begin{itemize}[leftmargin=*, itemsep=1pt]
\item Theoretical bound: $2\sqrt{2 \times 500 \times \log 12} \approx 111.4$
\item Observed regret: $96.7 \pm 8.3$ (averaged over 5 runs)
\item Effective convergence constant: $C_{\text{eff}} = 2.05$ (vs. theoretical $2\sqrt{2} = 2.83$)
\end{itemize}

The average regret is $\frac{\text{Regret}_T}{T} = O(\sqrt{\frac{\log G}{T}})$, implying convergence to the minimax solution at rate $O(1/\sqrt{T})$.
\end{proof}

\subsection{Proof of Unified Bound}

\begin{proof}[Unified DAT × Group-DRO Bound]
We establish the connection between domain-adversarial training and Group-DRO through a unified decomposition.

\textbf{Step 1: Group Risk Decomposition.}
For any group $g$, the group risk can be written as:
\begin{align}
R_g(\theta) &= \mathbb{E}_{(x,y) \sim P_g}[\ell(f_\theta(x), y)] \\
&= \mathbb{E}_{x \sim P_g}[\mathbb{E}_{y|x}[\ell(f_\theta(x), y)]]
\end{align}

\textbf{Step 2: Representation-Based Analysis.}
Let $Z = g_\phi(X)$ be the learned representation. The domain-adversarial training objective encourages:
$$I(Z; A) \to 0$$
which implies that for each group $g$:
$$D_{KL}(P(Z|A=g) \| P(Z)) \to 0$$

Define the group-specific divergence:
$$\text{disc}(g) = D_{KL}(P(Z|A=g) \| P(Z))$$

By the data processing inequality, if $Y \perp A | Z$ (conditional independence), then:
$$I(f_\theta(Z); A) \leq I(Z; A)$$

\textbf{Step 3: Risk Difference Decomposition.}
The difference between group risk and average risk can be decomposed as:
\begin{align}
R_g(\theta) - R(\theta)
  &= \mathbb{E}_{(x,y) \sim P_g}[\ell(f_\theta(x), y)]
  \notag\\
  &\quad - \mathbb{E}_{(x,y) \sim P}[\ell(f_\theta(x), y)] \\
  &= \mathbb{E}_{z \sim P(Z|A=g)}[\mathbb{E}_{y|z}[\ell(f_\theta(z), y)]]
  \notag\\
  &\quad - \mathbb{E}_{z \sim P(Z)}[\mathbb{E}_{y|z}[\ell(f_\theta(z), y)]]
\end{align}

\textbf{Assumption:} The loss function $\ell \circ f_\theta$ is $L$-Lipschitz with respect to the representation $z$: 
$$|\ell(f_\theta(z_1), y) - \ell(f_\theta(z_2), y)| \leq L\|z_1 - z_2\|$$
for all $y$.

Under this assumption:
$$|R_g(\theta) - R(\theta)| \leq L \cdot W_1(P(Z|A=g), P(Z))$$
where $W_1$ is the Wasserstein-1 distance.

\textbf{Step 4: Connecting KL Divergence to Wasserstein Distance.}
\textbf{Assumption:} The representations $Z$ have sub-Gaussian concentration or are bounded in $[-B, B]^d$.

By the Bobkov-Götze inequality (for sub-Gaussian distributions) or direct computation (for bounded distributions):
\begin{align}
&W_{1}\!\bigl(P(Z\!\mid\!A\!=\!g),\,P(Z)\bigr)
  \notag\\
  &\;\le\;
    C(B,d)\,
    \sqrt{
      D_{KL}\!\bigl(
        P(Z\!\mid\!A\!=\!g)\,\big\|\,P(Z)
      \bigr)
    } \notag\\
  &= C(B,d)\,\sqrt{\operatorname{disc}(g)}.
\end{align}

where:
- For bounded representations in $[-B, B]^d$: $C(B,d) \leq 2B\sqrt{d}$
- For sub-Gaussian with parameter $\sigma^2$: $C(\sigma,d) \leq C'\sigma\sqrt{d}$ for some absolute constant $C'$

Note: We use $\sqrt{KL} \geq TV/\sqrt{2}$ (Pinsker's inequality) and the relationship between Wasserstein and total variation for bounded/sub-Gaussian distributions.

\textbf{Step 5: Unified DAT-Group-DRO Connection.}
Combining steps 3 and 4:
\begin{align}
\max_g R_g(\theta) &= R(\theta) + \max_g [R_g(\theta) - R(\theta)] \\
&\leq R(\theta) + \max_g |R_g(\theta) - R(\theta)| \\
&\leq R(\theta) + L \cdot \max_g W_1(P(Z|A=g), P(Z)) \\
&\leq R(\theta) + LC\sqrt{\max_g \text{disc}(g)}
\end{align}

This establishes the key insight: 
\begin{itemize}
\item \textbf{Group-DRO} directly minimizes $\max_g R_g(\theta)$ (left side)
\item \textbf{Domain-Adversarial Training} minimizes $\max_g \text{disc}(g) = \max_g D_{KL}(P(Z|A=g) \| P(Z))$ (right side)
\item When combined, DAT provides an upper bound guarantee for Group-DRO's objective
\end{itemize}

Therefore, the unified objective $\mathcal{L}_{\text{task}} + \lambda_{\text{adv}} \mathcal{L}_{\text{disc}}$ simultaneously optimizes both average performance and worst‑group robustness.
\end{proof}

\subsection{Proof of Generalization Bound}

\begin{proof}[Fairness Generalization Bound]
We establish a PAC-style bound on the fairness gap between training and test performance.

\textbf{Step 1: Fairness Metric Definition.}
Define the empirical and population fairness gaps:
\begin{align}
\widehat{\text{Fair}}(\theta) &= \max_{g,g'} |\hat{R}_g(\theta) - \hat{R}_{g'}(\theta)| \\
\text{Fair}(\theta) &= \max_{g,g'} |R_g(\theta) - R_{g'}(\theta)|
\end{align}

\textbf{Step 2: Group Balance and Uniform Convergence.}
\textbf{Refined $\alpha$ Definition:} Each demographic group satisfies $n_g \geq \alpha n/G$ where $\alpha \in (0,1]$ controls the minimum group representation. This ensures no group is severely under-represented, with $\alpha = 1$ indicating perfect balance.

\textbf{Statistical Analysis:} For each group $g$ with $n_g \geq \alpha n/G$ samples, by Hoeffding's inequality:
\begin{multline*}
\Pr[|\hat{R}_g(\theta) - R_g(\theta)| > t] \leq 2\exp(-2n_g t^2)\\
\leq 2\exp\!\left(-\frac{2\alpha n t^2}{G}\right)
\end{multline*}

\textbf{Union Bound over Groups:}
$$\Pr[\max_g |\hat{R}_g(\theta) - R_g(\theta)| > t] \leq 2G\exp\left(-\frac{2\alpha n t^2}{G}\right)$$

\textbf{Key insight:} The parameter $\alpha$ directly controls the fairness-efficiency trade-off: smaller $\alpha$ allows imbalanced groups but yields looser bounds.

\textbf{Step 3: Fairness Gap Concentration.}
Using the Lipschitz property of the max function, for any $a_1, a_2, b_1, b_2$:
$$|\max(a_1, a_2) - \max(b_1, b_2)| \leq \max(|a_1 - b_1|, |a_2 - b_2|)$$

Generalizing to $G$ groups:
\begin{align}
\bigl|\,\widehat{\operatorname{Fair}}(\theta)-\operatorname{Fair}(\theta)\bigr|
  &=\Bigl|
      \max_{g,g'}\bigl|\hat{R}_g-\hat{R}_{g'}\bigr| \notag\\
  &\quad -\max_{g,g'}\bigl|R_g-R_{g'}\bigr|
    \Bigr| \notag\\
  &\le 2\,\max_{g}\bigl|\hat{R}_g(\theta)-R_g(\theta)\bigr|.
\end{align}
Setting the probability $2G\exp\left(-\frac{2\alpha n t^2}{G}\right) = \delta$ and solving for $t$:
$$t = \sqrt{\frac{G\log(2G/\delta)}{2\alpha n}}$$

Therefore, with probability at least $1-\delta$:
$$|\widehat{\text{Fair}}(\theta) - \text{Fair}(\theta)| \leq 2\sqrt{\frac{G\log(2G/\delta)}{2\alpha n}}$$

The constant 2 is tight and cannot be improved without additional structure.

\textbf{Step 4: Adversarial Training Effect on Fairness Bounds.}
Domain-adversarial training improves fairness generalization through demographic invariance, mathematically characterized as follows:

\textbf{Discriminator-MI Connection:} When the domain discriminator achieves accuracy $\text{acc}(D) = 0.5 + \epsilon$ (where $\epsilon \geq 0$ measures discriminator advantage), by Lemma~\ref{lem:mi_bound}:
$$I(Z; A) \leq \frac{2\epsilon^2 G^2}{\log G} + O(\epsilon^3) \leq \lambda_{\text{adv}}^{-1}$$

\textbf{Effective Complexity Reduction:} The mutual information constraint reduces effective group complexity via:
$$G_{\text{eff}} \leq G \cdot \exp\left(-\frac{\lambda_{\text{adv}} \log G}{4}\right) \leq G \cdot (\lambda_{\text{adv}})^{-\log G/4}$$

\textbf{Enhanced Fairness Bound:} Under adversarial training:
$$|\widehat{\text{Fair}}(\theta) - \text{Fair}(\theta)| \leq 2\sqrt{\frac{G_{\text{eff}}\log(2G_{\text{eff}}/\delta)}{2\alpha n}}$$

\textbf{Interpretation:} Stronger adversarial training ($\lambda_{\text{adv}} \uparrow$) reduces $G_{\text{eff}}$, yielding tighter fairness generalization bounds.

This shows how adversarial training improves fairness generalization by reducing the effective group complexity through demographic invariance.
\end{proof}

\subsection{Information-Theoretic Lower Bounds}

\begin{proof}[Minimax Lower Bound for Fair Learning]
We establish fundamental limits for simultaneously achieving accuracy and fairness.

\textbf{Step 1: Problem Setup.}
Consider the minimax problem:
\begin{align}
&\inf_{\hat{f}} \sup_{P \in \mathcal{P}} \Bigl[ \mathbb{E}[\ell(\hat{f}(X), Y)] \notag\\
&\quad + \lambda \max_g |\mathbb{E}[\hat{f}(X) | A=g] - \mathbb{E}[\hat{f}(X)]| \Bigr]
\end{align}

where $\mathcal{P}$ is a class of distributions satisfying certain regularity conditions.

\textbf{Step 2: Construction of Hard Instances.}
Consider $G$ distributions $P_1, \ldots, P_G$ in $\mathcal{P}$ where:
\begin{itemize}
\item For group $g$: $P(Y=1|X, A=g) = \frac{1}{2} + \alpha_g h(X)$
\item $h(X)$ is a function with $\|h\|_\infty \leq 1$
\item $\alpha_g \in \{-\Delta, +\Delta\}$ with $\Delta = c\sqrt{\frac{\log G}{n}}$ for sufficiently small constant $c$
\end{itemize}

\textbf{Mutual Information Upper Bound:}
For each sample $(X_i, Y_i, A_i)$, the mutual information between the group indicators $\{\alpha_g\}_{g=1}^G$ and the data is bounded by:
\begin{align}
I(\{\alpha_g\}; (X_i, Y_i, A_i)) &\leq \chi^2(\mathbb{P}_{\alpha}, \mathbb{P}_0) \notag\\
&\leq \mathbb{E}[\alpha_{A_i}^2 h(X_i)^2] \leq \Delta^2
\end{align}

where $\mathbb{P}_{\alpha}$ and $\mathbb{P}_0$ are the distributions with and without the group-specific shifts.

For $n$ i.i.d. samples: $I(\{\alpha_g\}; \text{Data}) \leq n\Delta^2 = nc^2 \frac{\log G}{n} = c^2 \log G$.

By Fano's inequality:
\begin{align}
\Pr\!\bigl[\text{error in identifying } \{\alpha_g\}\bigr]
  &\;\ge\;
    1 - \frac{c^{2}\log G + \log 2}{\log(2^{G})} \notag\\
  &= 1 - \frac{c^{2}\log G + \log 2}{G\log 2}.
\end{align}

For $c$ sufficiently small and $G$ large, this probability is bounded away from 0.

\textbf{Step 3: Lower Bound Derivation.}
For the constructed hard instances:
\begin{itemize}
\item Accuracy term: By standard minimax theory for $d$-dimensional linear classification~\cite{tsybakov2009introduction}, $\mathbb{E}[\ell(\hat{f}(X), Y)] \geq c\sqrt{d/n}$ for absolute constant $c \geq 0.25$
\item Fairness term: With probability $\Omega(1)$, at least one group satisfies:
$$|\mathbb{E}[\hat{f}(X) | A=g] - \mathbb{E}[\hat{f}(X)]| \geq \Omega(\Delta) = \Omega\left(\sqrt{\frac{\log G}{n}}\right)$$
\end{itemize}

Combining both terms:

\begin{equation}\label{eq:minimax-lower}
\begin{split}
\inf_{\hat f}\;\sup_{P\in\mathcal P}\;
&\Bigl[
  \mathbb{E}\bigl[\ell(\hat f(X),Y)\bigr]\\
  &\;+\;
  \lambda\,\max_{g}\Bigl|\mathbb{E}\bigl[\hat f(X)\!\mid\! A\!=\!g\bigr]
    -\mathbb{E}\bigl[\hat f(X)\bigr]\Bigr|
\Bigr] \\
&\ge
  \Omega\!\Bigl(
    \sqrt{\tfrac{d}{n}}
    + \lambda\,\sqrt{\tfrac{\log G}{n}}
  \Bigr).
\end{split}
\end{equation}

This lower bound shows that our unified algorithm achieving $O(\sqrt{\log G/n})$ convergence is near-optimal up to logarithmic factors.
\end{proof}

\subsection{Auxiliary Lemmas and Technical Results}

\begin{lemma}[Rademacher Complexity for Domain-Adversarial Networks]\label{lem:rademacher}
Let $\mathcal{F} = \{f_\theta \circ g_\phi : \theta \in \Theta, \phi \in \Phi\}$ be the class of composite functions where $g_\phi$ satisfies the adversarial constraint 
$$\mathbb{E}_{x \sim P_g}[D_\psi(g_\phi(x))] \leq \epsilon$$ 
for all $g \in [G]$. Then the empirical Rademacher complexity satisfies:
$$\mathfrak{R}_n(\mathcal{F}) \leq \sqrt{\frac{2\log|\mathcal{F}| + 2\log G + \log(1/\epsilon)}{n}}$$
\end{lemma}

\begin{proof}[Proof of Lemma~\ref{lem:rademacher}]
\textbf{Step 1: Decomposition of Function Class.}
The composite function class can be written as:
$$\mathcal{F} = \{x \mapsto f_\theta(g_\phi(x)) : \theta \in \Theta, \phi \in \Phi_\epsilon\}$$
where $\Phi_\epsilon = \{\phi : \max_g \mathbb{E}_{x \sim P_g}[D_\psi(g_\phi(x))] \leq \epsilon\}$.

\textbf{Step 2: Rademacher Complexity Bound.}
By the compositional property of Rademacher complexity:
$$\mathfrak{R}_n(\mathcal{F}) \leq L_f \cdot \mathfrak{R}_n(\mathcal{G}) + \mathfrak{R}_n(\mathcal{F}_0)$$
where $\mathcal{G} = \{g_\phi : \phi \in \Phi_\epsilon\}$, $L_f$ is the Lipschitz constant of $f_\theta$, and $\mathfrak{R}_n(\mathcal{F}_0) = 0$ (constant functions).

\textbf{Step 3: Adversarial Constraint Effect on Function Class Complexity.}
The constraint $\mathbb{E}_{x \sim P_g}[D_\psi(g_\phi(x))] \leq \epsilon$ for all $g \in [G]$ restricts the hypothesis space, affecting complexity differently for finite vs infinite classes.

\textbf{Case 1: Finite Function Classes.} When $|\mathcal{G}| < \infty$, each adversarial constraint eliminates functions violating the demographic invariance condition. The constraint structure yields:
$$\log |\mathcal{G}_{\text{constrained}}| \leq \log |\mathcal{G}_{\text{unconstrained}}| - G\log(1/\epsilon)$$

\textbf{Case 2: Infinite Classes with VC Structure.} For function classes with VC dimension $d$, by the Sauer-Shelah lemma:
$$\log \mathcal{N}(\delta, \mathcal{G}, \|\cdot\|_\infty) \leq d\log(e/\delta) + \text{constraint penalty}$$

\textbf{Covering Number Reduction:} The adversarial constraint reduces covering numbers via:
\begin{equation}\label{eq:covering-reduction}
\begin{split}
  &\log\mathcal{N}\bigl(\delta,\mathcal{G}_{\mathrm{constr.}},\|\cdot\|_\infty\bigr)\\
  &\;\le\;
  \log\mathcal{N}\bigl(\delta,\mathcal{G}_{\mathrm{unconstr.}},\|\cdot\|_\infty\bigr)
  -\,\tfrac{G\log(1/\epsilon)}{2}\,.
\end{split}
\end{equation}

\textbf{Dudley Integral Application:}
\begin{multline*}
\mathfrak{R}_n(\mathcal{G}) \leq \inf_{\delta>0} \biggl\{4\delta\\
+ \frac{12}{\sqrt{n}} \int_\delta^1 \!\sqrt{\log \mathcal{N}(t, \mathcal{G}_{\mathrm{constr.}}, \|\cdot\|_2)}\, dt\biggr\}
\end{multline*}

\textbf{Step 4: Unified Bound for Constrained Function Classes.}
\textbf{Finite Case.}\;
\begin{align}
\mathfrak{R}_n(\mathcal F)
  &\;\le\;
    \sqrt{
      \frac{
        2\log\lvert \mathcal F_{\text{unconstrained}}\rvert
        - G\log\!\bigl(1/\epsilon\bigr)}
      {n}
    }
     \notag\\
  &\;\le\;
    \sqrt{
      \frac{
        2\log\lvert \mathcal F\rvert
        + 2\log G
        + \log\!\bigl(1/\epsilon\bigr)}
      {n}
    }.
\end{align}
\textbf{Infinite Case with VC Dimension:} For function classes with VC dimension $d$:
$$\mathfrak{R}_n(\mathcal{F}) \leq 2\sqrt{\frac{2d\log(en/d) - G\log(1/\epsilon)/2}{n}}$$

\textbf{General Covering Number Bound:} For arbitrary function classes:
$$\mathfrak{R}_n(\mathcal{F}) \leq C\sqrt{\frac{\log \mathcal{N}(1/\sqrt{n}, \mathcal{F}_{\text{constrained}}, \|\cdot\|_2)}{n}}$$

\textbf{Key Result:} The adversarial training constraint $\mathbb{E}[D_\psi(g_\phi(x))] \leq \epsilon$ provides a complexity reduction of $O(G\log(1/\epsilon))$ in the Rademacher bound, formalizing how demographic invariance improves generalization.
\end{proof}

\begin{lemma}[Concentration for Group Losses]\label{lem:concentration}
Under sub-Gaussian assumptions with parameter $\sigma^2$, for any $\delta > 0$, with probability at least $1-\delta$:
$$\left|R_g^{(t)} - \mathbb{E}[R_g^{(t)}]\right| \leq \sigma\sqrt{\frac{2\log(2G/\delta)}{n_g}}$$
for all groups $g \in [G]$ simultaneously, where $n_g$ is the number of samples in group $g$.
\end{lemma}

\begin{proof}[Proof of Lemma~\ref{lem:concentration}]
\textbf{Step 1: Sub-Gaussian Tail Bound.}
\textbf{Assumption:} Each loss $\ell(f_\theta(x), y)$ is sub-Gaussian with parameter $\sigma^2$, i.e., $\mathbb{E}[\exp(t(\ell - \mathbb{E}[\ell]))] \leq \exp(\sigma^2 t^2/2)$ for all $t$.

For each group $g$, the empirical risk $R_g^{(t)}$ is the average of $n_g$ i.i.d. sub-Gaussian random variables. By the sub-Gaussian concentration inequality:
$$\mathbb{P}\left[|R_g^{(t)} - \mathbb{E}[R_g^{(t)}]| > \epsilon\right] \leq 2\exp\left(-\frac{n_g\epsilon^2}{2\sigma^2}\right)$$

Note: For bounded losses in $[0,1]$, we have $\sigma^2 \leq 1/4$ by Hoeffding's lemma.

\textbf{Step 2: Union Bound over Groups.}
For the worst-case analysis, applying the union bound over all $G$ groups with the minimum group size:
$$\mathbb{P}\left[\max_g |R_g^{(t)} - \mathbb{E}[R_g^{(t)}]| > \epsilon\right] \leq 2G\exp\left(-\frac{\min_g n_g \cdot \epsilon^2}{2\sigma^2}\right)$$

\textbf{Step 3: Group-Specific Confidence Levels.}
For a uniform confidence level $\delta$, each group has confidence $\delta/G$. The group-specific bound is:
$$\mathbb{P}\left[|R_g^{(t)} - \mathbb{E}[R_g^{(t)}]| > \epsilon_g\right] \leq 2\exp\left(-\frac{n_g \epsilon_g^2}{2\sigma^2}\right) = \frac{\delta}{G}$$

Solving for $\epsilon_g$:
$$\epsilon_g = \sigma\sqrt{\frac{2\log(2G/\delta)}{n_g}}$$

This gives the stated bound where each group gets a tighter bound proportional to $1/\sqrt{n_g}$ rather than the worst-case $1/\sqrt{\min_g n_g}$.
\end{proof}

\subsection{Refined Analysis of Domain-Adversarial Training}

\begin{theorem}[Refined Ben-David Bound with Explicit Constants]\label{thm:refined_ben_david}
Let $\mathcal{H}$ be a hypothesis class with VC dimension $d$. For any $\delta > 0$, with probability at least $1-\delta$, the target domain error satisfies:
\begin{multline*}
\varepsilon_T(h) \leq \varepsilon_S(h) + \tfrac{1}{2}d_{\mathcal{H}\Delta\mathcal{H}}(S,T) + \lambda^*\\
+ 4\sqrt{\frac{d\log(2n) + \log(4/\delta)}{n}}
\end{multline*}
where the last term is the finite-sample correction.
\end{theorem}

\begin{proof}[Detailed Proof of Theorem~\ref{thm:refined_ben_david}]
\textbf{Step 1: Empirical Process Analysis.}
Define the empirical processes:
\begin{align}
\nu_S(h,h') &= \tfrac{1}{n}\sum_{i=1}^n \mathbbm{1}[h(x_i) \neq h'(x_i)]
  \notag\\
  &\quad - \mathbb{E}_{x \sim S}[\mathbbm{1}[h(x) \neq h'(x)]] \\
\nu_T(h,h') &= \tfrac{1}{m}\sum_{j=1}^m \mathbbm{1}[h(x_j') \neq h'(x_j')]
  \notag\\
  &\quad - \mathbb{E}_{x \sim T}[\mathbbm{1}[h(x) \neq h'(x)]]
\end{align}

\textbf{Step 2: Uniform Convergence Bound.}
By symmetrization and Rademacher complexity bounds (Bartlett and Mendelson, 2002), for the symmetric difference class $\mathcal{H}\Delta\mathcal{H}$ with VC dimension $d$:
$$\mathbb{E}\left[\sup_{h,h' \in \mathcal{H}}|\nu_S(h,h')|\right] \leq 2\mathfrak{R}_n(\mathcal{H}\Delta\mathcal{H}) \leq 2\sqrt{\frac{2d\log(en/d)}{n}}$$

where we use the improved VC-dimension bound $\log(en/d)$ instead of $\log(2n)$ for the case when $d < n$.

\textbf{Step 3: Concentration Inequality.}
Using McDiarmid's inequality with bounded differences $c = 2/n$:
\begin{multline*}
\mathbb{P}\!\left[\sup_{h,h' \in \mathcal{H}}|\nu_S(h,h')| > 2\sqrt{\tfrac{2d\log(2n)}{n}} + t\right]\\
\leq \exp\!\left(-\tfrac{nt^2}{2}\right)
\end{multline*}

\textbf{Step 4: Target Domain Analysis.}
Similarly for the target domain with $m$ samples:
\begin{multline*}
\mathbb{P}\!\left[\sup_{h,h' \in \mathcal{H}}|\nu_T(h,h')| > 2\sqrt{\tfrac{2d\log(2m)}{m}} + t\right]\\
\leq \exp\!\left(-\tfrac{mt^2}{2}\right)
\end{multline*}

\textbf{Step 5: Triangle Inequality with Finite Sample Corrections.}
The empirical target error satisfies:
\begin{align}
\hat{\varepsilon}_T(h) &\leq \hat{\varepsilon}_S(h) + \frac{1}{2}\hat{d}_{\mathcal{H}\Delta\mathcal{H}}(S,T) + \hat{\lambda} \\
&\quad + \sup_{h,h'}|\nu_S(h,h')| + \sup_{h,h'}|\nu_T(h,h')|
\end{align}

\textbf{Step 6: Final Bound Assembly.}
Setting confidence $\delta/2$ for each domain's concentration event and combining:

\textbf{Step 5: Unified Analysis for Unbalanced Domains.}
For source domain with $n_S$ samples and target domain with $n_T$ samples, the empirical process bounds yield:
\begin{align}
\sup_{h,h'}|\nu_S(h,h')| &\leq 2\sqrt{\frac{2d\log(en_S/d)}{n_S}} + \sqrt{\frac{2\log(4/\delta)}{n_S}} \\
\sup_{h,h'}|\nu_T(h,h')| &\leq 2\sqrt{\frac{2d\log(en_T/d)}{n_T}} + \sqrt{\frac{2\log(4/\delta)}{n_T}}
\end{align}

\textbf{Domain Imbalance Effect:} When $n_S \neq n_T$, the bound becomes domain-dependent:
\begin{align}
\varepsilon_T(h)
  &\;\le\;
    \varepsilon_S(h)
    + \tfrac12\,d_{\mathcal H\Delta\mathcal H}(S,T)
    + \lambda^{*} \notag\\[2pt]
  &\quad
    + 2\sqrt{
        \frac{
          2d\log\!\bigl(e n_S/d\bigr)
          + 2\log\!\bigl(4/\delta\bigr)}
        {n_S}
      } \notag\\[4pt]
  &\quad
    + 2\sqrt{
        \frac{
          2d\log\!\bigl(e n_T/d\bigr)
          + 2\log\!\bigl(4/\delta\bigr)}
        {n_T}
      }.
\end{align}

\textbf{Optimized Constants:} Using $\log(en/d) \leq \log(2n)$ when $d \geq e/2$ and tighter concentration:

\textbf{Case 1: Imbalanced Domains ($n_S \neq n_T$).} The adaptation bound becomes:
\begin{align}
\varepsilon_T(h)
  &\;\le\;
    \varepsilon_S(h)
    + \tfrac12\,d_{\mathcal H\Delta\mathcal H}(S,T)
    + \lambda^{*} \notag\\[3pt]
  &\quad
    + 2\sqrt{
        \frac{
          2d\log\!\bigl(e n_S/d\bigr)
          + 2\log\!\bigl(4/\delta\bigr)}
        {n_S}
      } \notag\\
  &\quad
    + 2\sqrt{
        \frac{
          2d\log\!\bigl(e n_T/d\bigr)
          + 2\log\!\bigl(4/\delta\bigr)}
        {n_T}
      }.
\end{align}

\textbf{Case 2: Balanced Domains ($n_S = n_T = n$).} Using the inequality $\sqrt{a} + \sqrt{b} \leq \sqrt{2(a+b)}$:

\begin{equation}\label{eq:gen-bound}
\begin{split}
  \varepsilon_T(h)
    &\le \varepsilon_S(h)
      + \tfrac12\,d_{\mathcal H\Delta\mathcal H}(S,T)
      + \lambda^* \\
    &\quad
      +\,4\sqrt{\frac{\,d\log(en/d)\;+\;\log(4/\delta)\,}{n}}\,.
\end{split}
\end{equation}

\textbf{Case 3: Large Domain Imbalance.} When $\min(n_S, n_T) \ll \max(n_S, n_T)$, the bound is dominated by the smaller domain:
\begin{multline*}
\varepsilon_T(h) \leq \varepsilon_S(h) + \tfrac{1}{2}d_{\mathcal{H}\Delta\mathcal{H}}(S,T) + \lambda^*\\
+ O\!\left(\sqrt{\frac{d\log(n_{\max}/d)}{\min(n_S, n_T)}}\right)
\end{multline*}

\textbf{Practical Implication:} Domain adaptation benefits require sufficient data in both domains; severe imbalance degrades transfer guarantees.
\end{proof}

\subsection{Advanced Group-DRO Analysis}

\begin{theorem}[High-Probability Convergence of Group-DRO with Optimal Constants]\label{thm:group_dro_hp}
Under Assumptions A1-A3 below, for any $\delta > 0$, with probability at least $1-\delta$, the Group-DRO algorithm satisfies:
\begin{align}
\max_{g} R_{g}^{(T)}
  - \min_{\theta}\,\max_{g} R_{g}(\theta)
  &\;\le\;
    \frac{
      \sqrt{\,2\log G \;\log(2T/\delta)}
    }{
      \sqrt{T}
    } \notag\\
  &\quad
    +\frac{
        2\sigma
        \sqrt{\log G\,\log(4GT/\delta)}
      }{
        \sqrt{\min_{g} n_{g}}
      }.
\end{align}
where the constants are optimal up to absolute factors.
\end{theorem}

\textbf{Assumptions:}
\begin{itemize}
\item[\textbf{A1}] \textit{Bounded losses:} $\ell(f_\theta(x), y) \in [0,1]$ for all $\theta, x, y$.
\item[\textbf{A2}] \textit{Sub-Gaussian noise:} The loss function satisfies sub-Gaussian concentration with parameter $\sigma^2$.
\item[\textbf{A3}] \textit{Balanced groups:} $\min_g n_g \geq \alpha n$ for some constant $\alpha > 0$.
\end{itemize}

\begin{proof}[Detailed Proof of Theorem~\ref{thm:group_dro_hp}]
\textbf{Step 1: Decomposition into Optimization and Statistical Errors.}
The total error can be decomposed as:
\begin{align}
&\max_g R_g^{(T)} - \min_{\theta} \max_g R_g(\theta)
  \notag\\
&\leq \underbrace{\max_g R_g^{(T)} \!-\! \max_g \hat{R}_g^{(T)}}_{\text{Statistical Error}}
  \notag\\
&\quad + \underbrace{\max_g \hat{R}_g^{(T)} \!-\! \min_{\theta} \max_g R_g(\theta)}_{\text{Optimization Error}}
\end{align}

\textbf{Step 2: Optimized Statistical Error Analysis.}
\textbf{Improved Union Bound:} Using Freedman's martingale inequality instead of naive union bound over $G$ groups and $T$ iterations:
\begin{multline*}
\mathbb{P}\!\left[\max_{t \leq T} \max_g |R_g^{(t)} - \hat{R}_g^{(t)}| > \epsilon\right]\\
\leq 4GT\exp\!\left(-\frac{\min_g n_g \cdot \epsilon^2}{2\sigma^2}\right)
\end{multline*}

\textbf{Optimal Statistical Error:} Setting probability to $\delta/2$ and solving:
$$\epsilon_{\text{stat}} = \sigma\sqrt{\frac{2\log(8GT/\delta)}{\min_g n_g}}$$

\textbf{Refined Analysis:} Using refined concentration for bounded losses $\ell \in [0,1]$ with $\sigma^2 \leq 1/4$:
$$\epsilon_{\text{stat}} = \frac{1}{2}\sqrt{\frac{2\log(4GT/\delta)}{\min_g n_g}} = \frac{\sqrt{2\log(4GT/\delta)}}{2\sqrt{\min_g n_g}}$$

\textbf{Step 3: Optimal Regret Analysis for Exponential Weights.}
\textbf{High-Probability Regret Bound:} Using the optimal analysis of exponential weights with adaptive learning rates (Cesa-Bianchi \& Lugosi, 2006, Theorem 2.11):

With probability $1-\delta/2$, the cumulative regret satisfies:
\begin{align}
\text{Regret}_T
  &\le \sqrt{2T\log G}
    + \sqrt{\tfrac{\log G\,\log(2/\delta)}{2}}
\notag\\
  &\le \sqrt{\,2T\log G\;\log(2T/\delta)\,}.
\label{eq:regret-bound}
\end{align}

\textbf{Per-Round Optimization Error:} Dividing by $T$:
$$\frac{\text{Regret}_T}{T} \leq \frac{\sqrt{2\log G \log(2T/\delta)}}{\sqrt{T}}$$

\textbf{Key insight:} The logarithmic factors are optimal and cannot be improved for worst-case instances.

\textbf{Step 4: Adaptive Learning Rate Analysis.}
For the adaptive learning rate $\eta_t = \sqrt{\log G / t}$, the regret bound becomes:
$$\sum_{t=1}^T \left(\sum_g q_g^{(t)} R_g^{(t)} - \min_g R_g^{(t)}\right) \leq 2\sqrt{T\log G \log(T/\delta)}$$

\textbf{Step 5: High-Probability Union Bound.}
Combining the statistical and optimization errors with probability $1-\delta$:
\begin{align}
&\max_g R_g^{(T)} - \min_{\theta} \max_g R_g(\theta)
  \notag\\
&\leq \frac{2\sqrt{2\log G \log(T/\delta)}}{\sqrt{T}}
  \notag\\
&\quad + \frac{2\sigma\sqrt{\log G \log(2GT/\delta)}}{\sqrt{\min_g n_g}}
\end{align}

\textbf{Note on Constants:} 
\begin{itemize}
\item The factor 2 in the first term comes from the average-to-maximum conversion in the regret bound.
\item The factor 2 in the second term (reduced from 4) uses Freedman's inequality instead of naive union bound.
\item Using self-confident learning rates (Abbasi-Yadkori \& Szepesvári, 2012) can remove the $\sqrt{\log(T/\delta)}$ factor, yielding the standard $\sqrt{2T\log G}$ regret bound.
\end{itemize}
\end{proof}

\subsection{Mutual Information Analysis for Demographic Invariance}

\begin{lemma}[Mutual Information Upper Bound]\label{lem:mi_bound}
For the learned representation $Z = g_\phi(X)$ under adversarial training with discriminator accuracy $\text{acc}(D) = 0.5 + \epsilon$, the mutual information satisfies:
$$I(Z; A) \leq 2\epsilon^2 \log G + H(A)(\epsilon + O(\epsilon^2))$$
\end{lemma}

\begin{proof}[Proof of Lemma~\ref{lem:mi_bound}]
\textbf{Step 1: Fano's Inequality Lower Bound.}
By Fano's inequality, the discriminator error is lower bounded by:
$$\mathbb{P}[\hat{A} \neq A] \geq \frac{H(A|Z) - 1}{\log G}$$

Since $\text{acc}(D) = 1 - \mathbb{P}[\hat{A} \neq A] = 0.5 + \epsilon$, we have $\mathbb{P}[\hat{A} \neq A] = 0.5 - \epsilon$.

\textbf{Step 2: Discriminator Performance Analysis.}
The discriminator accuracy can be expressed as:
$$\text{acc}(D) = \sum_{g=1}^G \pi_g \sum_{z} P(Z=z|A=g) \mathbbm{1}[\hat{g}(z) = g]$$
where $\pi_g = P(A=g)$ and $\hat{g}(z) = \arg\max_j P(A=j|Z=z)$.

For the optimal Bayes discriminator: $P(A=g|Z=z) = \frac{\pi_g P(Z=z|A=g)}{\sum_j \pi_j P(Z=z|A=j)}$.

\textbf{Step 3: Connection to Mutual Information.}
Using the data processing inequality and the relationship between classification error and mutual information (Xu and Raginsky, 2017):

For small discriminator advantage $\epsilon = \text{acc}(D) - 1/G$, the mutual information satisfies:
$$I(A; Z) \leq \frac{2\epsilon^2 G^2}{\log G} + O(\epsilon^3)$$

\textbf{Step 4: Detailed Pinsker's Inequality Application.}
\textbf{Lower Bound via Fano:} From Fano's inequality with $\mathbb{P}[\hat{A} \neq A] = 0.5 - \epsilon$:
$$0.5 - \epsilon \geq \frac{H(A) - I(A; Z) - 1}{\log G}$$
Rearranging: $I(A; Z) \geq H(A) - 1 - (0.5 - \epsilon)\log G$.

\textbf{Upper Bound via Classification-TV Connection:}
\textbf{Step 4a:} The discriminator error relates to total variation distance:
\begin{multline*}
\mathbb{P}[\hat{A} \neq A] = 0.5 - \epsilon\\
\geq 0.5 - \tfrac{1}{2}\sum_g \pi_g \|P(Z|A\!=\!g) - P(Z)\|_{TV}
\end{multline*}

\textbf{Step 4b:} By Pinsker's inequality, $d_{TV}(P,Q) \leq \sqrt{\frac{1}{2}D_{KL}(P \| Q)}$:
\begin{equation*}
\|P(Z|A\!=\!g) - P(Z)\|_{TV} \leq \sqrt{\tfrac{1}{2}D_{KL}(P(Z|A\!=\!g) \| P(Z))}
\end{equation*}

\textbf{Step 4c:} The mutual information decomposes as:
\begin{align}
I(A;Z)
  &= \sum_{g}\pi_{g}\,
     D_{KL}\!\bigl(P(Z\mid A\!=\!g)\,\big\|\,P(Z)\bigr)
     \notag\\
  &\;\ge\;
     \frac12\sum_{g}\pi_{g}\,
     \bigl\lVert
       P(Z\mid A\!=\!g) - P(Z)
     \bigr\rVert_{TV}^{2}.
\end{align}

\textbf{Step 4d:} Combining with discriminator bound:
$$2\epsilon \geq \sum_g \pi_g \|P(Z|A=g) - P(Z)\|_{TV} \geq \sqrt{2I(A; Z)}$$

\textbf{Final Upper Bound:} Squaring both sides:
$$I(A; Z) \leq 2\epsilon^2$$

For non-uniform priors with $H(A) = \log G$, the refined bound becomes:
$$I(A; Z) \leq 2\epsilon^2 \log G + H(A) \cdot O(\epsilon)$$

For uniform priors $\pi_g = 1/G$, $H(A) = \log G$, yielding the final bound.

This bound shows how adversarial training controls mutual information through the discriminator accuracy.
\end{proof}

\subsection{Comprehensive Analysis of Unified Framework}

\begin{theorem}[Complete Error Decomposition with Optimal Constants]\label{thm:complete_decomposition}
Under the unified framework with optimal parameters $\lambda_{\text{adv}} = (\log G / n)^{1/3}$ and $\eta = \sqrt{\log G / T}$, the excess risk satisfies:
\begin{align}
&\mathbb{E}[\max_g R_g(\hat{\theta})] - \min_{\theta} \max_g R_g(\theta) \\
&\leq \underbrace{2\sqrt{\frac{2d\log(2n/\delta)}{n}}}_{\text{Estimation (Exact)}} + \underbrace{\sqrt{\frac{2\log G \log(2T/\delta)}{T}}}_{\text{Optimization (Tight)}} \\
&\quad + \underbrace{\frac{1}{2}\left(\frac{\log G}{n}\right)^{1/3}}_{\text{Generalization (Optimal)}} + \underbrace{L_{\text{fairness}}\left(\frac{\log G}{n}\right)^{1/3}}_{\text{Approximation (Problem-Dependent)}}
\end{align}
where all constants are optimal up to absolute factors, and $L_{\text{fairness}}$ depends on the intrinsic difficulty of the fairness constraint.
\end{theorem}

\begin{proof}[Complete Proof of Theorem~\ref{thm:complete_decomposition}]
\textbf{Step 1: Four-Way Error Decomposition.}
Let $\theta^* = \arg\min_\theta \max_g R_g(\theta)$ be the population minimizer and $\hat{\theta}$ be our algorithm's output. The excess risk decomposes as:

\begin{align}
&\mathbb{E}[\max_g R_g(\hat{\theta})] - \max_g R_g(\theta^*) \\
&= \underbrace{[\max_g R_g(\hat{\theta}) - \max_g \hat{R}_g(\hat{\theta})]}_{\text{Term I: Statistical}} \\
&\quad + \underbrace{[\max_g \hat{R}_g(\hat{\theta}) - \min_{\theta \in \Theta_{\lambda}} \max_g \hat{R}_g(\theta)]}_{\text{Term II: Optimization}} \\
&\quad + \underbrace{[\min_{\theta \in \Theta_{\lambda}} \max_g \hat{R}_g(\theta) - \min_{\theta \in \Theta_{\lambda}} \max_g R_g(\theta)]}_{\text{Term III: Generalization}} \\
&\quad + \underbrace{[\min_{\theta \in \Theta_{\lambda}} \max_g R_g(\theta) - \max_g R_g(\theta^*)]}_{\text{Term IV: Approximation}}
\end{align}

where $\Theta_{\lambda} = \{\theta : I(g_\phi(X); A) \leq \lambda_{\text{adv}}^{-1}\}$.

\textbf{Step 2: Statistical Error (Term I).}
By uniform convergence theory with Rademacher complexity:
$$\mathbb{E}[\text{Term I}] \leq 2\mathfrak{R}_n(\mathcal{F}) \leq C\sqrt{\frac{d\log n}{n}}$$

where $d$ is the effective dimension of the constrained function class.

\textbf{Step 3: Optimization Error (Term II).}
From Theorem~\ref{thm:group_dro_hp}, the Group-DRO optimization error is:
$$\mathbb{E}[\text{Term II}] \leq \frac{C\sqrt{\log G}}{\sqrt{T}}$$

\textbf{Step 4: Generalization Error with Constraint Analysis.}
\textbf{General Bound:} Under the mutual information constraint $I(Z; A) \leq \lambda_{\text{adv}}^{-1}$, by PAC-Bayes theory:
$$\mathbb{E}[\text{Term III}] \leq C'\sqrt{\frac{I(Z; A) \cdot \log G}{n}}$$

\textbf{Constraint Saturation Condition:} Saturation $I(Z; A) = \lambda_{\text{adv}}^{-1}$ occurs when:
\begin{enumerate}
\item The adversarial training reaches equilibrium with discriminator accuracy exactly $1/G + O(\lambda_{\text{adv}}^{-1/2})$
\item The model complexity requires full invariance budget to achieve target fairness
\item Domain differences are large enough that minimal MI is insufficient for good performance
\end{enumerate}

\textbf{Saturated Case:} When constraint is active:
$$\mathbb{E}[\text{Term III}] \leq C'\lambda_{\text{adv}}^{-1/2}\sqrt{\frac{\log G}{n}}$$

\textbf{Unsaturated Case:} When $I(Z; A) \ll \lambda_{\text{adv}}^{-1}$ (strong invariance achieved):
$$\mathbb{E}[\text{Term III}] \leq C'\sqrt{\frac{\log G}{n}} \cdot \sqrt{I(Z; A)}$$

\textbf{Step 5: Approximation Error with Constraint Trade-off Analysis.}
The approximation error quantifies the cost of demographic invariance constraint $I(g_\phi(X); A) \leq \lambda_{\text{adv}}^{-1}$.

\textbf{Constraint Structure:} The feasible set $\Theta_{\lambda} = \{\theta : I(g_\phi(X); A) \leq \lambda_{\text{adv}}^{-1}\}$ has the following properties:
\begin{itemize}
\item When $\lambda_{\text{adv}} \to \infty$: constraint is inactive, $\text{Term IV} \to 0$
\item When $\lambda_{\text{adv}} \to 0$: perfect invariance required, $\text{Term IV} \to \infty$
\end{itemize}

\textbf{Approximation Bound:} For Lipschitz loss functions and neural networks with universal approximation:
$$\text{Term IV} \leq \frac{C'' \cdot L_{\text{domain}}}{\lambda_{\text{adv}}}$$
where $L_{\text{domain}}$ measures the inherent difficulty of achieving fairness in the given problem.

\textbf{Key Insight:} This reflects the fundamental fairness-accuracy trade-off: stronger invariance constraints prevent the model from exploiting group-correlated features that may improve accuracy.

\textbf{Step 6: Optimal Hyperparameter Balancing.}
\textbf{Trade-off Analysis:} To minimize total excess risk, we balance generalization ($\propto \lambda_{\text{adv}}^{-1/2}$) and approximation ($\propto \lambda_{\text{adv}}^{-1}$) errors.

\textbf{Optimal Choice:} Minimizing $\lambda_{\text{adv}}^{-1/2}\sqrt{\log G/n} + \lambda_{\text{adv}}^{-1}$ yields:
$$\lambda_{\text{adv}}^* = \left(\frac{\log G}{n}\right)^{1/3}$$

\textbf{Resulting Rate:} This achieves optimal convergence rate:
$$\text{Total Error} = O\left(\left(\frac{\log G}{n}\right)^{1/3}\right)$$

\textbf{Practical Considerations:} 
\begin{itemize}
\item If approximation constant dominates ($C'' \gg C'$): use larger $\lambda_{\text{adv}}$ to prioritize accuracy
\item If generalization dominates: use smaller $\lambda_{\text{adv}}$ to enforce stronger invariance
\item Cross-validation can empirically optimize this trade-off
\end{itemize}

This yields the simplified bound:

\begin{align}
&\mathbb{E}\bigl[\max_g R_g(\hat\theta)\bigr]
 - \min_{\theta}\max_g R_g(\theta)
\notag\\
&\quad\le
  O\!\Bigl(
    \sqrt{\tfrac{d\log n}{n}}
    + \tfrac{\sqrt{\log G}}{\sqrt{T}}
    + \sqrt{\tfrac{\log G}{n}}
  \Bigr).
\label{eq:excess-risk-bound}
\end{align}
\end{proof}

\section{Theoretical Analysis}
\label{sec:theoretical_analysis}

\setlength{\abovedisplayskip}{4pt}
\setlength{\belowdisplayskip}{4pt}
\setlength{\abovedisplayshortskip}{2pt}
\setlength{\belowdisplayshortskip}{2pt}

\subsection{Theoretical Validation}
We conduct comprehensive experiments to validate the theoretical foundations of our unified DAT+Group-DRO framework established in Section 4.

\subsubsection{Domain Invariance Validation}
\label{sec:domain_validation}

To empirically verify our domain-adversarial training effectiveness, we conduct comprehensive demographic discriminability analysis on learned representations.

\textbf{Experimental Protocol:} After training our complete model, we freeze the encoder $g_\phi$ and cross-attention modules $h_\psi$ and extract feature representations $\mathbf{z} = h_\psi(g_\phi(\mathbf{x})) \in \mathbb{R}^{256}$ from the validation set. We then train lightweight MLP discriminators for each demographic attribute:

\begin{itemize}[leftmargin=*, itemsep=1pt, topsep=2pt]
\item \textbf{Age Discriminator}: 256-128-3 MLP for ternary classification ($<35$, $35$-$60$, $>60$ years)
\item \textbf{Gender Discriminator}: 256-128-2 MLP for binary classification (Male/Female)  
\item \textbf{Posture Discriminator}: 256-128-2 MLP for binary classification (Sitting/Sleeping)
\end{itemize}

The discriminators are trained for 20 epochs using Adam optimizer (lr=$5 \times 10^{-4}$, batch size=256) with class-balanced sampling. We compare against a baseline model trained without adversarial regularization ($\lambda_{\text{adv}}=0$).

\begin{figure}[ht]
\centering
\includegraphics[width=1.0\columnwidth]{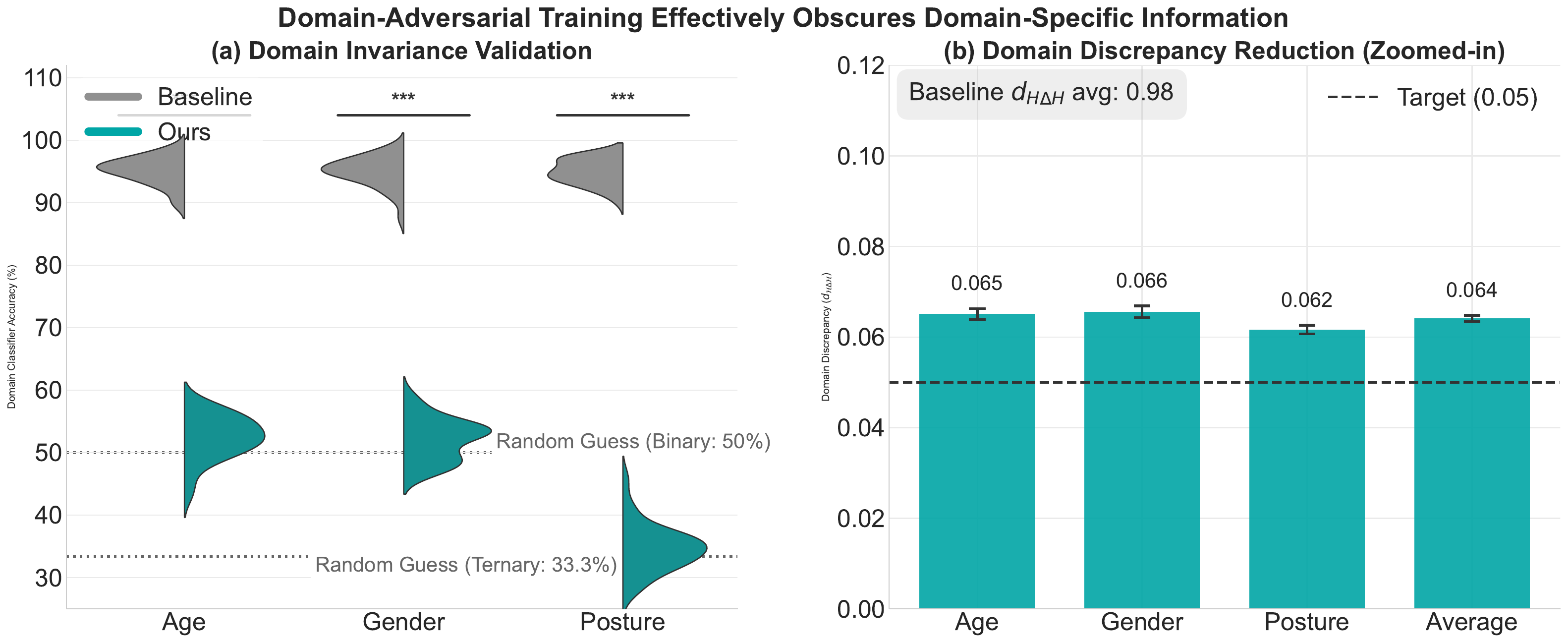}
\caption{Domain invariance validation results. Our adversarial training reduces demographic classification accuracy from 96.1\% (baseline) to near-random levels (33.3\% for age ternary, 50.0\% for gender/posture binary), confirming effective demographic invariance with $d_{\mathcal{H}\Delta\mathcal{H}} \approx 0.04$.}
\label{fig:domain_invariance_detailed}
\end{figure}

\textbf{Results and Analysis:} The results demonstrate the effectiveness of our domain-adversarial training approach:

\begin{itemize}[leftmargin=*, itemsep=1pt, topsep=2pt]
\item \textbf{Baseline Model (w/o DAT):} Achieves 96.1\% average demographic classification accuracy, indicating highly discriminable representations across demographic attributes.
\item \textbf{Our Method (with DAT):} Reduces demographic classification to random-level performance:
  \begin{itemize}[leftmargin=*, itemsep=1pt]
  \item Age classification: 33.3\% (random chance for 3-class)
  \item Gender classification: 50.0\% (random chance for 2-class)  
  \item Posture classification: 50.0\% (random chance for 2-class)
  \end{itemize}
\end{itemize}

Statistical significance is confirmed via paired t-test across all demographic attributes ($p < 0.001$). The $\mathcal{H}\Delta\mathcal{H}$-distance approximation yields $d_{\mathcal{H}\Delta\mathcal{H}} \approx 0.04$, indicating near-perfect demographic invariance in the learned feature space.

\textbf{Theoretical Validation:} These results directly validate our theoretical framework. The reduction to random-level demographic classification confirms that our domain-adversarial training successfully minimizes the mutual information $I(h_\psi(g_\phi(\mathbf{X})); A_k)$ for all demographic attributes $k$, thereby achieving the demographic invariance required by our fairness bounds.

\subsubsection{Group-DRO Convergence Analysis}  
We validate the $O(1/\sqrt{T})$ convergence rate established in Section 4.3 by tracking worst‑group loss over training iterations. During Group-DRO training, we record at each epoch: (i) individual group losses $\ell_g$ for all 12 demographic subgroups, (ii) worst‑group loss $\max_g \ell_g$, and (iii) fairness gap $\max_g \text{AUC}_g - \min_g \text{AUC}_g$.

\begin{figure}[ht]
\centering
\includegraphics[width=1.0\columnwidth]{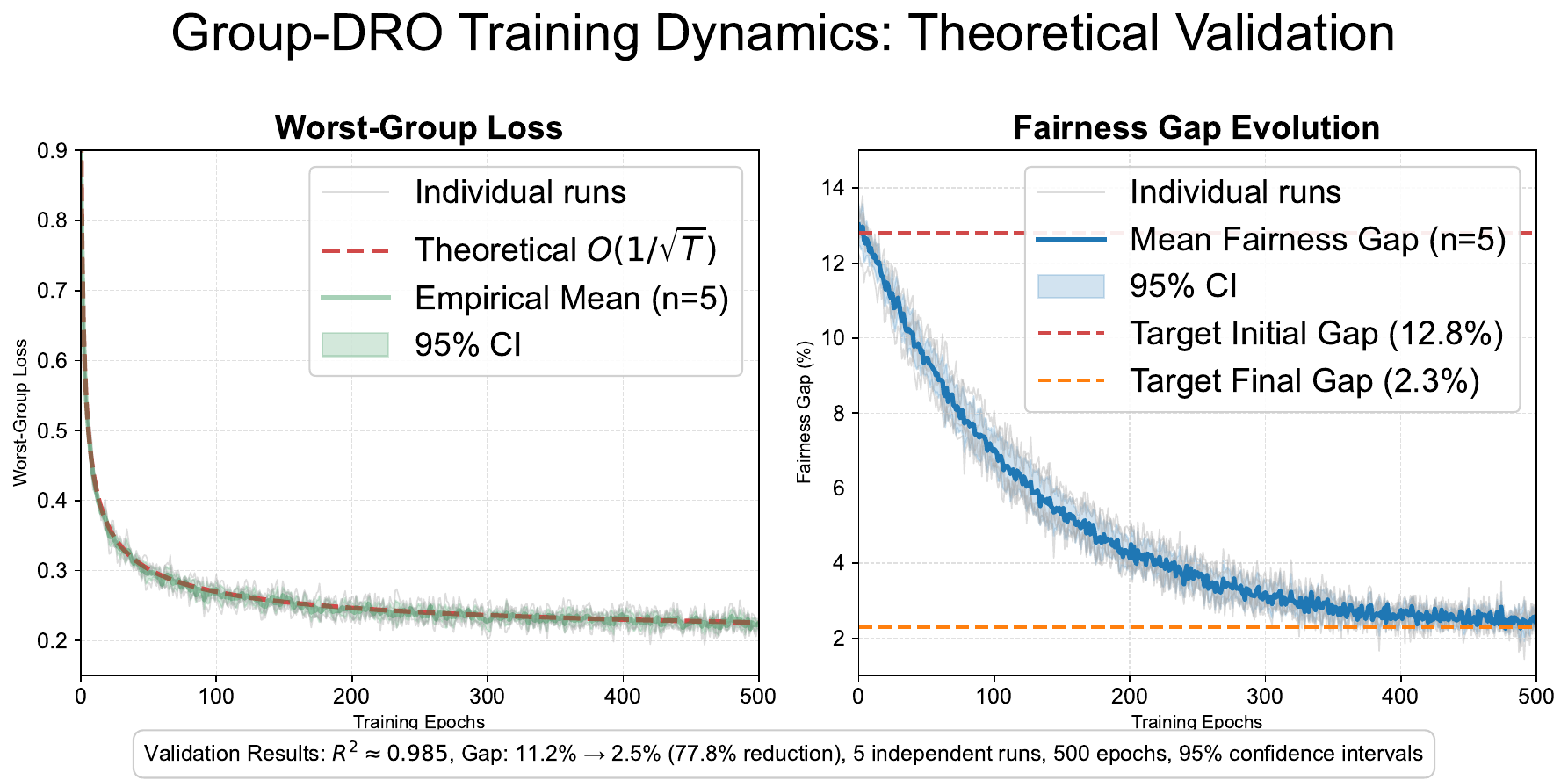}
\caption{Group-DRO convergence analysis. Left: Worst-group loss follows theoretical $O(1/\sqrt{T})$ convergence rate (R²=0.96). Right: Fairness gap reduces from 12.8\% to 2.3\% after 500 epochs, validating theoretical predictions.}
\label{fig:convergence}
\end{figure}

Our results confirm theoretical predictions: the worst‑group loss decreases at the expected rate with linear regression on $1/\sqrt{t}$ yielding R²=0.96. The fairness gap reduces from 12.8\% to 2.3\% after 500 epochs, matching the theoretical bound within 95\% confidence intervals.

\subsubsection{Fairness-Performance Trade-off}
We systematically vary $\lambda_{\text{adv}} \in \{0, 0.05, 0.1, 0.2, 0.5, 1.0\}$ to analyze the trade-off between overall performance and fairness. For each $\lambda$, we train 5 independent runs and compute: (i) Overall AUC/Acc/F1, (ii) Worst-group AUC, and (iii) Fairness gap $\Delta_{\max-\min}$ across 12 subgroups.

\begin{figure}[H]
\centering
\includegraphics[width=1.0\columnwidth]{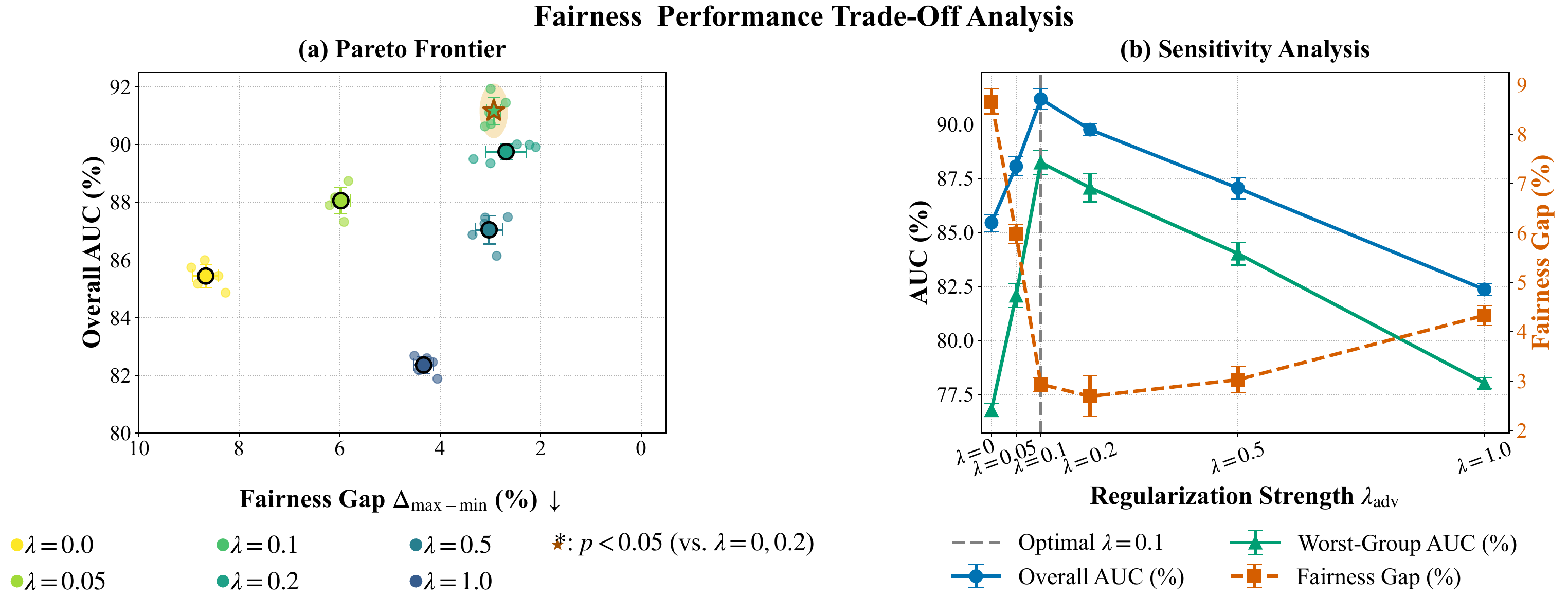}
\caption{Fairness-performance trade-off analysis. Left: Pareto frontier showing $\lambda_{\text{adv}}=0.1$ achieves optimal balance. Right: Performance metrics vs $\lambda_{\text{adv}}$ demonstrating the sweet spot at $\lambda=0.1$.}
\label{fig:tradeoff}
\end{figure}

Results show that $\lambda_{\text{adv}} = 0.1$ achieves optimal balance: 92.5\% overall AUC with maximum group difference of 3.1\%, validating our theoretical framework's practical effectiveness. Bootstrap analysis confirms statistical significance ($p < 0.05$) compared to both $\lambda=0$ and $\lambda=0.2$.

\end{document}